\documentclass[10pt,journal,compsoc]{IEEEtran}

\usepackage[pdftex]{graphicx}
\usepackage{times}
\usepackage{epsfig}
\usepackage{tabu}
\usepackage{amsmath}
\usepackage{amssymb}
\usepackage{array}
\usepackage{booktabs}
\usepackage{multirow}
\usepackage{float}
\usepackage{ctable}
\usepackage{xspace}
\usepackage[switch]{lineno}
\DeclareMathOperator{\atan}{atan}

\makeatletter
\newcommand{\thickhline}{%
    \noalign {\ifnum 0=`}\fi \hrule height 1pt
    \futurelet \reserved@a \@xhline
}
\newcolumntype{"}{@{\hskip\tabcolsep\vrule width 1pt\hskip\tabcolsep}}

\DeclareRobustCommand\onedot{\futurelet\@let@token\@onedot}
\def\@onedot{\ifx\@let@token.\else.\null\fi\xspace}

\def\eg{\emph{e.g}\onedot} 
\def\ie{\emph{i.e}\onedot} 
 
\def\etc{\emph{etc}\onedot} 
 
\def\etal{\emph{et al}\onedot}

\makeatother

\ifCLASSOPTIONcompsoc
  \usepackage[nocompress]{cite}
\else
  \usepackage{cite}
\fi

\begin{document}

%\linenumbers

\title{{Towards} Reading Hidden Emotions: A Comparative Study of Spontaneous Micro-expression Spotting and Recognition Methods}

\author{Xiaobai~Li,
           Xiaopeng~Hong,
           Antti~Moilanen, %~\IEEEmembership{Student Member,~IEEE,}
           Xiaohua~Huang,
           Tomas~Pfister, 
		Guoying~Zhao,
		and~Matti~Pietik\"{a}inen

\thanks{X. Li, X. Hong, A. Moilanen, X. Huang, G. Zhao and M. Pietik\"{a}inen are with the Center for Machine Vision and Signal Analysis, University of Oulu, Finland.}
\thanks{E-mail: \{xiaobai.li, xhong,  xiaohua.huang, gyzhao, mkp\}@ee.oulu.fi, antti.moilanen@aalto.fi.}
\thanks{T. Pfister is with Department of Engineering Science, University of Oxford, Oxford, UK. E-mail: tp@robots.ox.ac.uk}
\thanks{This research is supported by the Academy of Finland, Tekes and Infotech Oulu.}
\thanks{\textcopyright 2017 IEEE. Personal use of this material is permitted. Permission from IEEE must be obtained for all other uses, including reprinting/republishing this material for advertising or promotional purposes, collecting new collected works for resale or redistribution to servers or lists, or reuse of any copyrighted component of this work in other works.}}

\maketitle
\vspace{-4ex}
\begin{abstract}
Micro-expressions (MEs) are rapid, involuntary facial expressions which reveal emotions that people do not intend to show. 
Studying MEs is valuable as recognizing them has many important applications, particularly in forensic science and psychotherapy. 
However, analyzing spontaneous MEs is very challenging due to their short duration and low intensity.
Automatic ME analysis includes two tasks: ME spotting and ME recognition.
For ME spotting, previous studies have focused on posed rather than spontaneous videos.
For ME recognition, the performance of previous studies is low. 
To address these challenges, we make the following contributions:
(i)~We propose the first method for spotting spontaneous MEs in long videos (by exploiting feature difference contrast).
This method is training free and works on arbitrary unseen videos.
(ii)~We present an advanced ME recognition framework, which outperforms previous work by a large margin on two challenging spontaneous ME databases (SMIC and CASMEII).
(iii)~We propose the first automatic ME analysis system (MESR), which can spot and recognize MEs from spontaneous video data.
Finally, we show our method outperforms humans in the ME recognition task by a large margin, and achieves comparable performance to humans at the very challenging task of spotting and then recognizing spontaneous MEs.
\end{abstract}


\vspace{-2ex}
%%%%%%%%%%%%%%%%%%%%%%%%%%%%%%%%%%%%%%%%%%%%%%%%%%%%%%%
\section{Introduction}
\IEEEPARstart{F}{acial} expressions (FE) are one of the major ways that humans convey emotions. Aside from ordinary FEs that we see every day, under certain circumstances emotions can also manifest themselves in the special form of micro-expressions (ME). 
{An ME is a very brief, involuntary FE shown on people's face according to experienced emotions.
ME may occur in high-stake situations when people try to conceal or mask their true feelings for either gaining advantage or avoiding loss\cite{ekman2003darwin}. }
In contrast to ordinary FEs, 
MEs are very short (1/25 to 1/3 second, the precise length definition varies \cite{yan2013fast,matsumoto2011evidence}), and 
the intensities of involved muscle movements are subtle~\cite{porter2008reading}.

The phenomenon was first discovered by Haggard and Isaacs \cite{haggard1966micromomentary} in 1966, who called them ‘micromomentary facial expressions’. Three years later, Ekman and Friesen also reported finding MEs~\cite{ekman1969nonverbal} when they were examining a video of a psychiatric patient, trying to find possible trait of her suicide tendency. Although the patient seemed happy throughout the film, a fleeting look of anguish lasting two frames ($1/12$s) was found when the tape was examined in slow motion. This feeling of anguish was soon confirmed through a confession from the patient in her another counseling session: she lied to conceal her plan to commit suicide. In the following decades, Ekman and his colleague continued researching MEs~\cite{%ekman1991can,
frank1997ability,ekman2003darwin,ekman2002microexpression}. Their work has drawn increasing interests from both academic and commercial communities.

A major reason for the considerable interest in MEs is that it is an important clue for lie detection~\cite{ekman2009lie,ekman2003darwin}.
Spontaneous MEs occur fast and involuntarily, and they are difficult to control through one's willpower \cite{ekman2007emotions}. 
In high-stake situations \cite{frank1997ability} for example when suspects are being interrogated, an ME fleeting across the face could give away a criminal pretending to be innocent, as the face is telling a different story than his statements. 
Furthermore, as has been demonstrated in \cite{ekman2009lie, ekman1999few}, people who perform better at ME recognition tests are also better lie detectors. 
Due to this, MEs are used as an important clue by police officers for lie detection in interrogations. 
Ekman developed a Micro Expression Training Tool (METT) \cite{ekman2002microexpression} to help improve the ME recognition abilities of law enforcement officers. 
In addition to law enforcement, ME analysis has potential applications in other fields as well. 
In psychotherapy, MEs may be used for understanding genuine emotions of the patients when additional reassurance is needed. 
{In the future when this technology becomes more mature, it might also be used to help} border control agents to detect abnormal behavior, and thus to screen potentially dangerous individuals during routine interviews.

However, detecting and recognizing MEs are very difficult for human beings~\cite{ekman1999few} (in contrast to normal FEs, which we can recognize effortlessly).  
Study~\cite{ekman2002microexpression} shows that for ME recognition tasks, people without training only perform slightly better than chance on average.
This is because MEs are too short and subtle for human eyes to process.
The performance can be improved with special training, but it is still far below the `efficient' level. 
Moreover, finding and training specialists to analyze these MEs is very time-consuming and expensive. 

Meanwhile, in computer vision, algorithms have been reported to achieve performance of above 90\% for ordinary FE recognition tasks on several FE databases \cite{zeng2009survey}.
It is reasonable to explore the use of computer algorithms for automatically analyzing MEs. 
FE recognition has become popular since Picard proposed the concept of Affective computing~\cite{picard1997book} in 1997. 
But studies of ME have been very rare until now due to several challenges. 
One challenge is the lack of available databases. 
It is difficult to gather spontaneous ME samples as MEs only occur under certain conditions, and the incidence rate is low. 
Other challenges include developing methods able to deal with the short durations and the low intensity levels of MEs. 

For the analysis of spontaneous ME there are two major tasks: 
first, \textit{spotting} when the ME occurs from its video context (ME Spotting); and
second, \textit{recognizing} what {expression} the ME represents (ME Recognition). 
In natural circumstances an ME may occur along with other more prominent motion such as head movements and eye blinks. This makes spotting and recognizing rapid and faint MEs very challenging.

To address these issues in spotting and recognizing spontaneous MEs, we make the following \textit{main contributions}:
(1)~We propose a new method for spotting spontaneous MEs. 
To our best knowledge, this is the first ME spotting method which {is demonstrated to be effective on spontaneous ME databases.} 
Our method is based on feature difference (FD) contrast and peak detection, and it is training free. 
An initial version of this work was published in~\cite{moilanen2014spotting}.
(2)~We develop an advanced framework for ME recognition. 
This new framework achieves much better performance than previous works because: 
first, we employed Eulerian video magnification method for motion magnification to counter the the low intensity of MEs; and 
secondly we investigated three different feature descriptors (Local Binary Pattern (LBP), Histograms of Oriented Gradients (HOG) and Histograms of Image Gradient Orientation (HIGO)) and their combinations on three orthogonal planes for this task. 
(3)~We provide comprehensive explorations of several key issues related to ME recognition. 
In particular, we draw the following conclusions basing on substantial experimental results: 
(i)~temporal interpolation (TIM) is valuable for ME recognition as it unifies the ME sequence lengths; 
(ii)~combining feature histograms on all three orthogonal planes (TOP) does \emph{not} consistently yield the best performance, as the XY plane may contain redundant information; 
(iii)~gradient-based features (HOG and HIGO) outperform LBP for ordinary color videos, while LBP features are more suitable for near-infrared (NIR) videos; and
(iv)~using a high speed camera facilitates ME recognition and improves results.
(4)~Lastly, we propose an automatic ME analysis system (MESR), which first spots and then recognizes MEs from long videos. 
In experiments on the SMIC database, we show that it achieves comparable performance to human subjects.

The remaining parts of this paper are organized as follows: 
Section \uppercase\expandafter{\romannumeral2} reviews related work; Section \uppercase\expandafter{\romannumeral3} introduces our ME spotting method; Section \uppercase\expandafter{\romannumeral4} describes our new ME recognition framework; and Section \uppercase\expandafter{\romannumeral5} presents experimental results and discussions for both ME spotting and recognition. Conclusions are drawn in Section \uppercase\expandafter{\romannumeral6}.
\vspace{-2ex}
%%%%%%%%%%%%%%%%%%%%%%%%%%%%%%%%%%%%%%%%%%%%%%%%%%
\section{Related Work}

\subsection{ME databases}

Adequate training data is a prerequisite for the development of ME analysis. 
Several \emph{FE} databases exist, such as JAFFE~\cite{kamachi1998japanese}, CK~\cite{kanade2000comprehensive}, MMI~\cite{pantic2005web} and others. 
These databases have enhanced the progress of algorithms for conventional FE analysis. 
However, when it comes to the available ME databases, the options are much more limited.

Eliciting spontaneous MEs is difficult. 
According to Ekman, an ME may occur under `high-stake' conditions~\cite{frank1997ability}, which indicate situations when an individual tries to hide true feelings (because (s)he knows that the consequences for being caught may be enormous). 
{In \cite{frank1997ability} Ekman \etal proposed} three ways to construct a high-stake situation: 
(1) asking people to lie about what they saw in videos; 
(2) constructing crime scenarios; and 
(3) asking people to lie about their own opinions. 
Different ways were used to motivate participants to lie successfully: for example, participants were informed that assessments of their performance (from professionals) will impact their future career development; or good liars got extra money as rewards. 
Maintaining these scenarios are demanding for the conductors of the experiments, and the occurrence of MEs is still low even under these conditions. 
Not everyone could yield an ME~\cite{li2013spontaneous,yan2013casme,yan2014casme}. 
The second challenge comes after data collection: labeling MEs is very challenging and time-consuming even for trained specialists. 

In some earlier studies on automatic ME analysis, posed ME data were used to bypass the difficulty of getting spontaneous data. 
Shreve~\etal~\cite{shreve2009towards,shreve2011macro} reported collecting a database called USF-HD which contains 100 clips of posed MEs. 
They first showed some example videos that contain MEs to the subjects, and then asked them to mimic those examples. 
There are limited details about the collected ME samples in their paper, and it is unclear what emotion categories were included or how the ground truth were labeled. 
The USE-HD data were used by the authors for automatic ME spotting using an optical flow method. 
Polikovsky~\etal~\cite{polikovsky2009facial} collected a posed ME database by asking subjects to perform seven basic emotions with low intensity and go back to neutral expression as quickly as possible. 
Ten subjects were enrolled and the posed MEs were recorded by a high speed camera (200 fps). 
Their data were labeled with AUs for each frame following the facial action coding system (FACS~\cite{ekman1978facial}).
The authors used a 3D-gradient orientation histogram descriptor for AU recognition. 

The biggest problem of posed MEs is that they are different from real, naturally occurring spontaneous MEs. 
Studies show~\cite{ekman1969nonverbal} that spontaneous MEs occur involuntarily, and that the producers of the MEs usually do not even realize that they have presented such an emotion.
By asking people to perform expressions as quickly as possible one can obtain posed MEs, but they are different from spontaneous ones in both spatial and temporal properties~\cite{porter2008reading,yan2013fast}. 
So methods trained on posed MEs can not really solve the problem of automatic ME analysis in practice.

Therefore it is important to obtain a large, representative spontaneous database of MEs. 
To this end, we collected the first spontaneous ME database SMIC~\cite{pfister2011recognising,li2013spontaneous}. 
In order to simulate a `high-stake' situation as Ekman suggested, we designed our ME inducing protocol as follows: 
we used video clips which were demonstrated to be strong emotion inducing materials in previous psychological studies~\cite{coan2007handbook, gross1995emotion} to make sure that subjects will have strong emotional reactions; 
we asked the subjects to hide their true feeling and keep a neutral face while watching the video clips, and report their true feelings afterwards. 
The high-stake situation was created by informing the subjects that they were being monitored through a camera while watching movies, and that they will be punished (filling a very long boring questionnaire) once we spot any emotion leaks on their faces. 

By using the above paradigm, we successfully induced spontaneous MEs from 16 out of 20 subjects.
The first version of SMIC includes 77 MEs from six subjects~\cite{pfister2011recognising}, and it was later extended to the full version of SMIC which includes 164 MEs from 16 subjects~\cite{li2013spontaneous}. 
The full version of SMIC contains three datasets (all with resolution of $640\times480$): 
(1)~an HS dataset recorded by a high speed camera at 100 fps, 
(2)~a VIS dataset recorded by a normal color camera at 25 fps; and 
(3)~an NIR dataset recorded by a near infrared camera both at 25 fps. 
The HS camera was used to record all data, while VIS and NIR cameras were only used for the recording of the last eight subjects' data. 
{The ME clips were segmented from the original long videos from onset to offset, and then labeled into three emotion classes: \textit{positive}, \textit{surprise} and \textit{negative}.
The assumption for the ME-emotion mapping is that: the occurred MEs reveal participants' true hidden emotions induced by the movies, so the labels should be consistent with both the content of the movie and with participants' own feedback.
Originally there were five emotion labels according to the contents of the movies, including \textit{happy}, \textit{sad}, \textit{disgust}, \textit{fear} and \textit{surprise} (anger movies were excluded because no ME of anger was induced during the prior test). The three labels of \textit{sad}, \textit{disgust} and \textit{fear} were later merged into one label of \textit{negative} because of two reasons, first is people may feel more than one of them at the same time thus make them indistinguishable according to the aforementioned labeling system; secondly the sample numbers will be better balanced this way.}
The labeling was performed by two annotators, first separately and then cross-checked. 
Only the labels that both annotators agreed on were included.

Yan and colleagues~\cite{yan2013casme} collected another spontaneous ME database using a similar emotion-inducing paradigm. 
The Chinese Academy of Sciences Micro-expression (CASME) database~\cite{yan2013casme} contains 195 MEs elicited from 19 participants. 
CASME data were recorded using two cameras: 
the BenQ M31 camera with frame rate of 60 fps and resolution of $1280\times720$ (CASME-A), and 
the Point Grey GRAS-03K2C camera with frame rate of 60 fps and resolution of $640\times480$ (CASME-B). 
One major difference between CASME and SMIC is that CASME has AU labels. 
Each ME clip was first labeled with AUs, and then classified into one of the eight emotion categories: amusement, sadness, disgust, surprise, contempt, fear, repression and tension.  

Later Yan~\etal~\cite{yan2014casme} collected a newer version of ME database, CASMEII, which provides more ME samples with higher spatial and temporal resolutions.
The new database includes 247 ME samples from 26 subjects.
The face videos were recorded at 200 fps, with an average face size of $280\times340$. 
The ME-inducing paradigm for CASMEII is similar to the one used in SMIC, and the data contains both AU labels and emotion labels of five classes (happiness, disgust, surprise, repression and other).
{The emotion labels were assigned based on AUs, participants' self reports, and the contents of the movies.}
\setlength{\floatsep}{0cm}
\vspace{-2ex}
\begin{table}[thbp]
\caption{Current micro-expression databases. Elicitation methods \textbf{P}/\textbf{S}: posed/spontaneous.}
\vspace{-1ex}
\centering
{\renewcommand{\arraystretch}{1.0}
\renewcommand{\tabcolsep}{0.05cm}
\begin{tabular}{|c|c|c|c|c|c|c|c|c|}
\hline
\multirow{2}{*}{\textbf{Database}} & {{USF-HD}} & {Polikovsky} & \multicolumn{3}{c|}{{SMIC \cite{li2013spontaneous}}} & \multicolumn{2}{c|}{{CASME \cite{yan2013casme}}} & {{CASMEII }}\\
\cline{4-8}
& \cite{shreve2011macro} & \cite{polikovsky2009facial} &{HS} & {VIS} & {NIR} & {A} & {B} & \cite{yan2014casme} \\
\hline\hline
\textbf{MEs} & 100 & N/A & 164 & 71 & 71 & 100 & 95 & 247\\
\hline
\textbf{Subjects} & N/A & 10 & 16 & \multicolumn{2}{c|}{8} & 7 & 12 & 26\\
\hline
\textbf{Fps} & 30 & 200 & 100 & 25 & 25 & \multicolumn{2}{c|}{60} & 200\\
\hline
\textbf{{Resolution}} & 720*1280 & 480*640 & \multicolumn{3}{c|}{640*480} & 1280*720 & 640*480 & 640*480\\
\hline
\textbf{Elicitation} & P & P & \multicolumn{3}{c|}{S} & \multicolumn{2}{c|}{S} & S\\
\hline
\textbf{Emotions} & N/A & 7 & \multicolumn{3}{c|}{3} & \multicolumn{2}{c|}{8} & 5\\
\hline
\end{tabular}}
\label{table:currentMEdatabases}
\end{table}

\begin{figure}[H]
\centering
\includegraphics[scale=0.5]{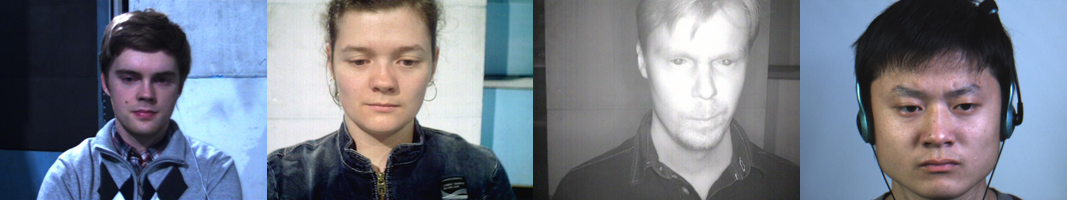}
\vspace{-2ex}
\caption{{Sample frames from SMIC and CASMEII databases. From left to right: SMIC-HS, SMIC-VIS, SMIC-NIR, CASMEII.}}
\label{fig:sampleFrames}
\end{figure}
\vspace{-2ex}
\setlength{\intextsep}{2pt}

Table~\ref{table:currentMEdatabases} lists the key features of existing ME databases. In this paper we focus on spontaneous MEs.
SMIC and CASMEII databases are used in our experiments as they are the most comprehensive spontaneous ME databases currently publicly available.
Sample frames from SMIC and CASMEII databases are shown in Figure~\ref{fig:sampleFrames}.

%%%%%%%%%%%%%%%%%%%%
\subsection{State of the art of ME spotting research}

ME spotting refers to the problem of automatically detecting the temporal interval of an ME in a sequence of video frames. 
Solutions to similar problems, such as spotting ordinary FEs, eye-blinking, and facial AUs from live videos~\cite{zeng2006spontaneous,krolak2012eye,%jeni2013continuous,
liwicki2013incremental} have been performed, but only a few studies have investigated automatic spotting of MEs.

Most existing works have tackled the problem in posed ME databases. 
{In papers \cite{polikovsky2009facial} and \cite{polikovsky2013facial}}, the authors used 3D gradient histograms as the descriptor to distinguish the onset, apex and offset stages of MEs from neutral faces. 
Although their method could potentially contribute to the problem of ME spotting, two drawbacks have to be considered.  
First, their method was only tested on posed MEs, which are much easier compared to spontaneous ME data. 
Second, their experiment was run as a classification test, in which all frames were classified into one of the four stages of MEs. 
{Classification requires sufficient amount of data and time for training before it can be used for spotting.}

Wu~\etal~\cite{wu2011machine} proposed to use Gabor filters to build an automatic ME recognition system which includes the spotting step. 
They achieved very high spotting performance on the METT training database~\cite{ekman2002microexpression}. 
However, the METT training samples are fully synthetic: they have been synthesized by inserting one FE image in the middle of a sequence of identical neutral face images.
This makes the spotting task significantly easier, as in real conditions the onset and offset of an ME would not be as abrupt, and the context frames would be more dynamic.  

Shreve~\etal~\cite{shreve2009towards,shreve2011macro} used an optical strain-based method to spot both macro (ordinary FEs -- the antonym for `micro') and micro expressions from videos. 
But as in the aforementioned studies, the data used in their experiments are posed. 
The authors reported being able to spot 77 from the 96 MEs, with 36 false positives on their posed database. 
The paper also evaluated their method on a small database of 28 spontaneous MEs found in TV or on-line videos, with results of 15 true positives and 18 false positives. But the database was small and not published.

As seen above, most previous ME spotting methods were tested only using posed data, which are different (and easier) compared to spontaneous MEs. 
In particular, during the recording of posed ME clips, subjects can voluntarily control their behavior according to the given instructions. 
Therefore one should expect the posed data to be more `clean', \eg contain restricted head movements, more clear-cut onsets and offsets, less movement in the context frames between two MEs, and so on. 
In contrast, the situation is more complicated in real videos which contain spontaneous MEs. 
This is because spontaneous MEs can appear during the presence of an ordinary FE (with either the same or the opposite valence of emotion~\cite{porter2008reading,warren2009detecting}), and sometimes they overlap with the occurrences of eye blinks, head movements and other motions. 
Therefore the spotting task is more challenging on spontaneous ME data. 
As an intermediate step to spot MEs in spontaneous data, several studies~\cite{pfister2011recognising,ruiz2013encoding,davison2014micro,yao2014micro} tackled an easier step which was referred to as ME `detection'.
ME detection is a two-class classification problem, in which a group of labeled ME clips are classified against the rest non-ME clips.
However, to our knowledge, none of these works present any experiments for spotting spontaneous MEs from long realistic videos.

\vspace{-2ex}
%%%%%%%%%%%%%%%%%%%%
\subsection{State of the art of ME recognition research}

ME recognition is the task where, given a `spotted' ME (\ie a temporal interval containing an ME), the ME is classified into two or more classes (\eg happy, sad \etc).
Experiments on this problem are more prominent in the literature, and have been carried out using both posed and spontaneous data.

Some researchers investigated ME recognition on posed ME databases. 
Polikovsky~\etal~\cite{polikovsky2009facial,polikovsky2013facial} used a 3D gradient descriptor for the recognition of AU-labeled MEs. 
Wu~\etal~\cite{wu2011machine} combined Gentleboost and an SVM classifier to recognize synthetic ME samples from the METT training tool. 
{It is worthy of noticing that AU detection is considered as an effective approach for ME recognition in previous works such as  \cite{polikovsky2009facial}. AU detection task was not explored in this paper because the SMIC database, which is the major testing database employed in our experiments, doesn't have AU labeling.}

Several recent studies also reported tests on spontaneous ME databases. 
{In \cite{pfister2011recognising} Pfister \etal were} the first to propose a spontaneous ME recognition method.
The method combined three steps: 
first, a temporal interpolation model (TIM) to temporally `expand' the micro-expression into more frames; 
second, LBP-TOP feature extraction; 
and third, using Multiple kernel learning for classification. 
The method achieved an accuracy of 71.4\% (two-class classification) for ME recognition on the first version of the SMIC database. 
In our later work~\cite{li2013spontaneous}, a similar method was tested on the full version of SMIC and the best recognition result was 52.11\% (three-class classification) on the VIS part (25fps RGB frames) of the database. 
Since then, LBP and its variants have often been employed as the feature descriptors for ME recognition in many other studies. 
Ruiz-Hernandez and Pietik{\"a}inen~\cite{ruiz2013encoding} used the re-parameterization of a second order Gaussian jet to generate more robust histograms, and achieved better ME recognition result than~\cite{pfister2011recognising} on the first version of SMIC database (six subjects).

Song~\etal~\cite{song2013learning} recognized emotions by learning a sparse codebook from facial and body micro-temporal motions. 
Although the concept of ME was employed in their study, their definition of MEs was wider than that of the current paper, as gestures from body parts (other than face) were also included. 
They ran experiments on a spontaneous audio-visual emotion database (AVEC2012), but not on any ME database. 
Wang~\etal~\cite{wang2014ICPR} extracted LBP-TOP from a Tensor Independent Colour Space (instead of ordinary RGB) for ME recognition, and tested their method on CASMEII database. 
In Wang \etal 's another paper \cite{wang2014micro}, Local Spatiotemporal Directional Features were used together with the sparse part of Robust PCA for ME recognition, achieving an accuracy of 65.4\% on CASMEII.

So far most ME recognition studies have considered using LBP-TOP as the feature descriptor.
As there is still much room for improvement in the recognition performance, more robust descriptors and machine learning methods need to be explored.

\vspace{-2ex}
%%%%%%%%%%%%%%%%%%%%%%%%%%%%%%%%%%%%%%%%%%%%%%%%%%%
{\section{Method for ME spotting}}

{In this section we present an ME spotting method that combines appearance-based feature descriptors with Feature Difference (FD) analysis. 
Our ME spotting method consists of four steps, as shown in Figure~\ref{fig:MEspottingDiagram}, and more details of each step are given below. }

\begin{figure}[H]
\centering
\includegraphics[scale=0.7]{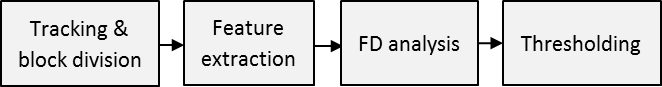}
\caption{{Work flow diagram for the proposed ME spotting method.}}
\label{fig:MEspottingDiagram}
\end{figure}
\setlength{\intextsep}{2pt}

\begin{figure}[H]
\centering
\includegraphics[scale=0.25]{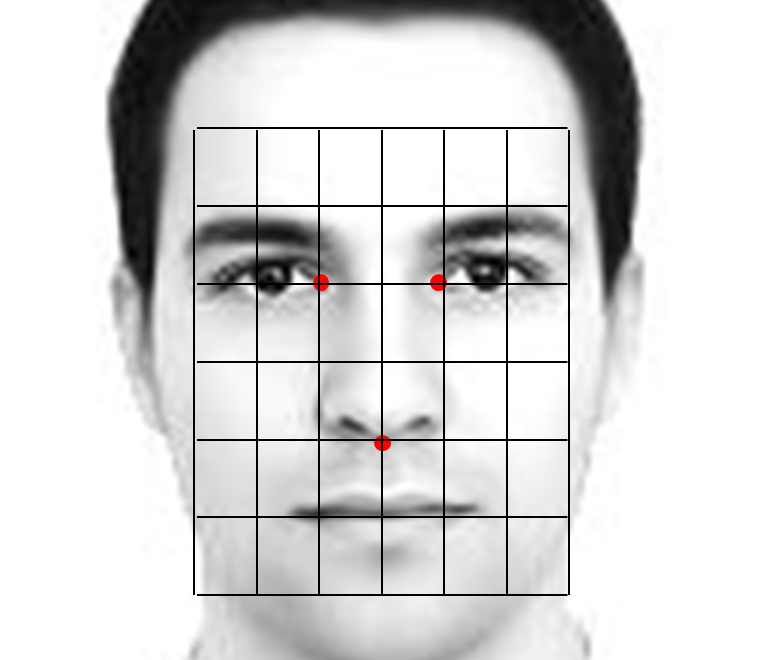}
\caption{{The face area is divided into $6 \times 6$ blocks according to the coordinates of three tracked facial feature points.}}
\label{fig:spottingBlocks}
\end{figure}

{\subsection{Facial points tracking and block division}}
{We detect two inner eye corners and a nasal spine point on the first frame and then track them through the sequence using the Kanade-Lucas-Tomasi algorithm~\cite{tomasi1991detection}. 
3D head rotation problem is not considered here since all faces in the employed testing databases are in near-frontal view.
The in-plane rotation and face size variations within the sequence are corrected based on the coordinates of the three tracked points. 
Then, the facial area is divided into equal-sized blocks. 
In order to keep the contents of each block still, the block structure is fixed according to the coordinates of the three points as shown in Figure~\ref{fig:spottingBlocks}. }

\setlength{\intextsep}{2pt}

{\subsection{Feature extraction}
Two feature descriptors are selected for evaluation: LBP~\cite{ojala2002multiresolution} and the Histogram of Optical Flow (HOOF)~\cite{liu2009beyond}. 
Over the past decade, LBP and its variants have been successfully applied to face recognition and ME recognition in several works~\cite{ahonen2006face,pfister2011recognising}. 
We first calculate normalized LBP histogram for each block, and then concatenate all the histograms to get the LBP feature vector for one frame.}

{Optical flow-based methods have lately gained popularity in FE research. 
For example, Shreve~\etal~\cite{shreve2009towards,shreve2011macro} used optical strain for micro- and macro-expression spotting on posed data. 
In our experiments, we also exploit this method and compare it to the LBP feature. 
In particular, we calculate Histogram of Optical Flow (HOOF) by obtaining the flow field for each frame (with one frame being the reference flow frame) and then compiling the orientations into a histogram. 
HOOF is implemented based on the code by Liu~\etal~\cite{liu2009beyond} with default parameters. 
Two kinds of reference frame are tested: one uses the first frame of the input video as the reference frame throughout the whole video; and the other uses the first frame within the current time window as the reference frame (and therefore changes as the time window slides through the video). We discuss the two options in the experiments.}

\vspace{-3ex}
\subsection{Feature difference (FD) analysis}
The FD analysis compares the differences of the appearance-based features of sequential video frames within a specified interval.
{First, we define several concepts for explaining the method. CF stands for current frame (the frame that is currently analyzed); when a micro-interval of $N$ frames is used, TF stands for the tail frame (the $k^{\text{th}}$ frame before the CF).
HF stands for the head frame (the $k^{\text{th}}$ frame after the CF). 
We define $k =1/2 \times (N-1)$.
The average feature frame (AFF) is defined as a feature vector representing the average of the features of TF and HF. }

\begin{figure}[H]
\centering
\includegraphics[scale=0.35]{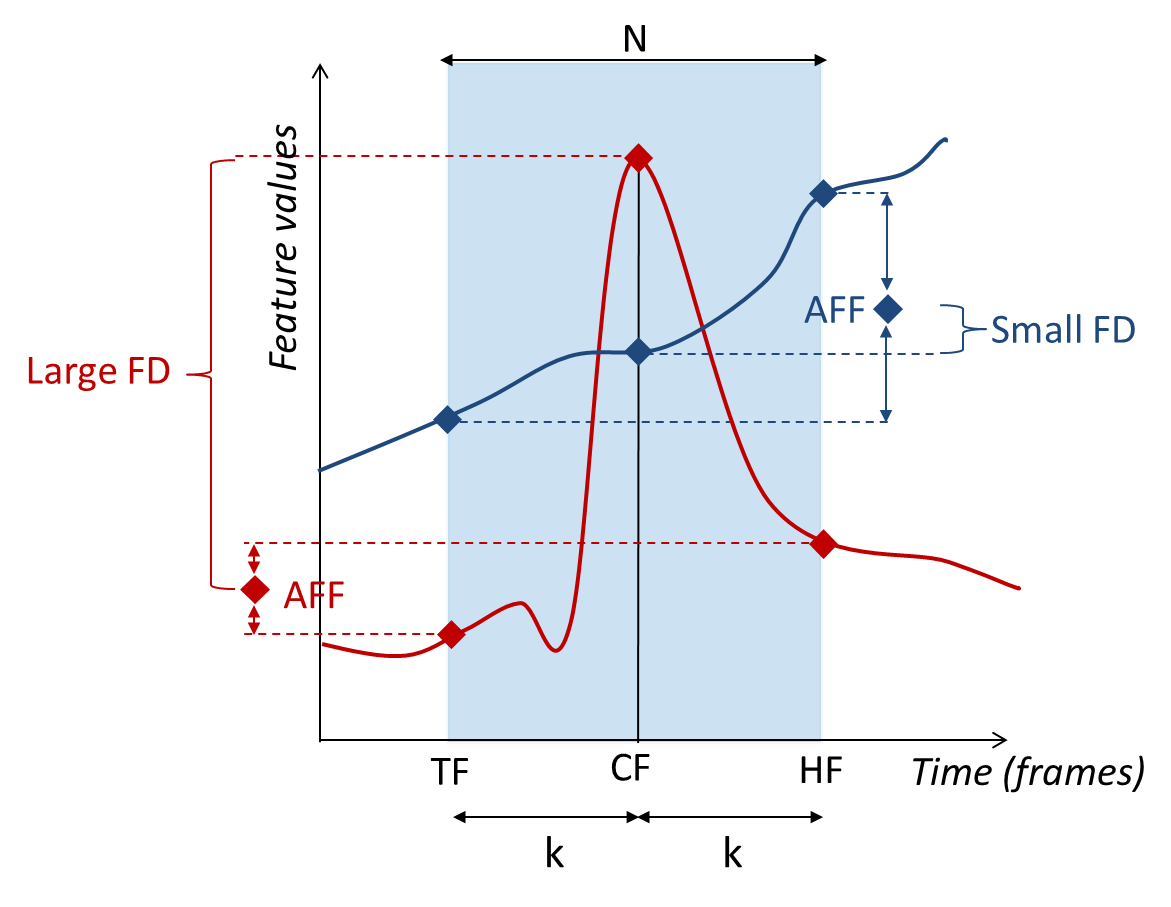}
\vspace{-2ex}
\caption{{Illustration of terms used in feature difference (FD) analysis. The red curve shows a rapid facial movement (\eg an ME) which produces a large FD; the blue curve shows a slower facial movement (\eg an ordinary FE) which produces smaller FD.}}
\label{fig:spottingFD}
\end{figure}

{The basic idea of FD analysis is as follows: for each CF, its features are compared to the respective AFF by calculating the dissimilarity of the feature vectors. 
By sliding a time window of $N$ frames, this comparison is repeated for each frame excluding the first $k$ frames from the beginning and the last $k$ frames at the end of the video, where TF and HF would exceed the video boundaries. 
Relevant terms are illustrated in Figure~\ref{fig:spottingFD}.}

\vspace{-2ex}
\setlength{\intextsep}{1pt}

{FD of a pair of feature histograms is calculated using the Chi-Squared (${\chi}^2$) distance. %~\cite{pele2010quadratic}
Large FD indicates a rapid facial movement (\eg an ME) with both an onset and offset phase occurring in the time window, as shown as the red curve in Figure~\ref{fig:spottingFD}.
In contrast, for slower movements with smoother onset and offset, the FD calculated within such a short time are significantly smaller, as the blue curve in Figure~\ref{fig:spottingFD}. }

\subsection{Thresholding and peak detection}

{For the $i$th frame of the input video, we calculate FD values for each of its 36 blocks as: $d_{i,1}, \dots, d_{i,36}$. The FD values are sorted in a descending order as $d_{i,j_{1}}, \dots, d_{i,j_{36}}$, where $j_{1}, \dots, j_{36} \in \{1,2, \dots, 36\}$.
The occurrence of an ME will result in larger FD values in some (but not all) blocks. We use the average of the $M$ greatest block FD values for each frame, and obtain an initial difference vector $F$ for the whole video as
\begin{equation}\label{eq:spot_F}
  F_i = \frac{1}{M}\sum^{M}_{\beta=1}d_{i,j_{\beta}},   
\end{equation}
where $i = 1, 2,\dots, n$, and $n$ is the frame number of the video. 
Here we use one third of all 36 blocks with the biggest FD values and set $M=12$. }

{To distinguish the relevant peaks from local magnitude variations and background noise, contrasting of the vector $F$ is done by subtracting the average of the surrounding TF and HF initial difference values from each CF value. 
Thus, the $i$th value in the contrasted difference vector becomes
\begin{equation}\label{eq:spot_C}
C_i=F_i-\frac{1}{2}(F_{i+k}+F_{i-k}),
\end{equation}
and the contrasted difference vector for the whole video is obtained by calculating $C$ for all frames except for the first and the last $k$ frames of the video. }

{After contrasting, all negative difference vector values are assigned to be zero as they indicate that there are no rapid changes of features in the CF comparing to that of TF and HF. 
Finally, threshold and peak detection are applied to locate the peaks indicating the highest intensity frames of rapid facial movements. 
The threshold $T$ is calculated as
\begin{equation}\label{eq:spot_T}
T=C_{\text{mean}}+\tau\times(C_{\text{max}}-C_{\text{mean}}),
\end{equation}
where $C_{\text{mean}}$ and $C_{\text{max}}$ are the average and the maximum of difference values for the whole video, and $\tau$ is a percentage parameter in the range of $[0, 1]$. 
Minimum peak distance in the peak detection is set to $k/2$. 
The spotted peaks will be compared with ground truth labels to tell whether they are true or false spots. 
Spotting results using different thresholds are presented and discussed in experiments.
More details about the method can be found in~\cite{moilanen2014spotting}. }

%%%%%%%%%%%%%%%%%%%%%%%%%%%%%%%%%%%%%%%%%%%%%%%%%%%%%%%
\section{Method for ME recognition}
\begin{figure}[H]
\centering
\includegraphics[scale=0.5]{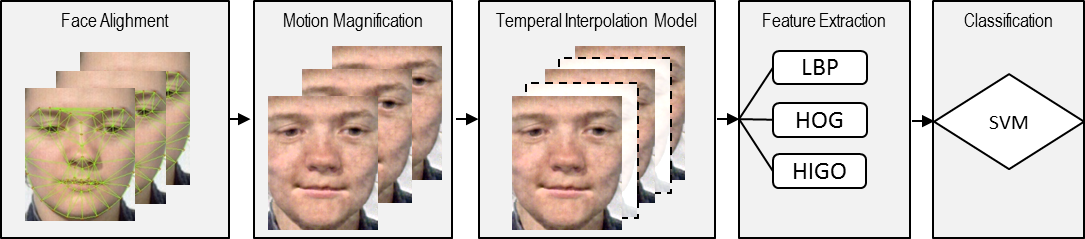}
\vspace{-2ex}
\caption{{Framework diagram for the proposed ME recognition method.}}
\vspace{-2ex}
\label{fig:rec_diagram}
\end{figure}

In this section we present the framework of our proposed method for ME recognition. 
An overview of our method is shown in Figure~\ref{fig:rec_diagram}.
The following subsections discuss the details of the method.

%%%%%%%%%%%%%%%%%%%%%
\subsection{Face alignment}

For ME spotting we do not align faces across different videos, because the method only compares frames within the current input video. 
However, for ME recognition, training is inevitable to deal with the large inter-class variation. 
Thus face alignment is needed to minimize the differences of face sizes and face shapes across different video samples. 

Let $V=\{v_1, v_2, \dots, v_l\}$ denote the whole set of ME clips from a testing database and $l$ is the total number of ME samples. 
The $i$th sample clip is given by {$v_i=(I_{i,1}, I_{i,2},\dots, I_{i,n_i})$,} where $I_{i,j}$ is the $j$th frame and $n_i$ is the frame number of the clip $v_i$. 

First, we select a frontal face image $I_{\text{mod}}$ with neutral expression as the model face. $68$ facial landmarks of the model face $\psi(I_{\text{mod}})$ are detected using the Active Shape Model~\cite{cootes1995active}. 

{Then the $68$ landmarks are detected on the first frame of each ME clip $I_{i,1}$. 
To further normalize the variations caused by different subjects and the movements, we rely on image transformation functions to build the correspondence between landmarks of $I_{i,1}$ and those of the model face $I_{\text{mod}}$ so that $I_{i,1}$ and remaining frames in a sequence can be registered to the model face. 
We choose the Local Weighted Mean (LWM)~\cite{goshtasby1988image} to compute a transform matrix for face registration, for its superiority in providing smooth transition across adjacent areas in a re-sampled image.}
The transform matrix $TRAN$ is:
\begin{equation}
TRAN_i=LWM(\psi(I_{\text{mod}}),\psi(I_{i,1})), \, i=1,\dots,l,
\end{equation}
where $\psi(I_{i,1})$ is the coordinates of $68$ landmarks of the first frame of the ME clip $v_i$. 
Then all frames of  $v_i$ were normalized using matrix $TRAN_i$. 
The normalized image $I'$ was computed as a 2D transformation of the original image:
\begin{equation}
I'_{i,j}=TRAN_i\times I_{i,j}, \, j=1,\dots,n_i,
\end{equation}
where $I'_{i,j}$ is the $j$th frame of the normalized ME clip $v'_i$. 

In the last step, we crop face areas out from normalized images of each ME clip using the rectangular defined according to the eye locations in the first frame $I'_{i,1}$.

%%%%%%%%%%%%%%%%%%%%%
\subsection{Motion magnification}

\begin{figure}[H]
\begin{center}
\includegraphics[scale=0.42]{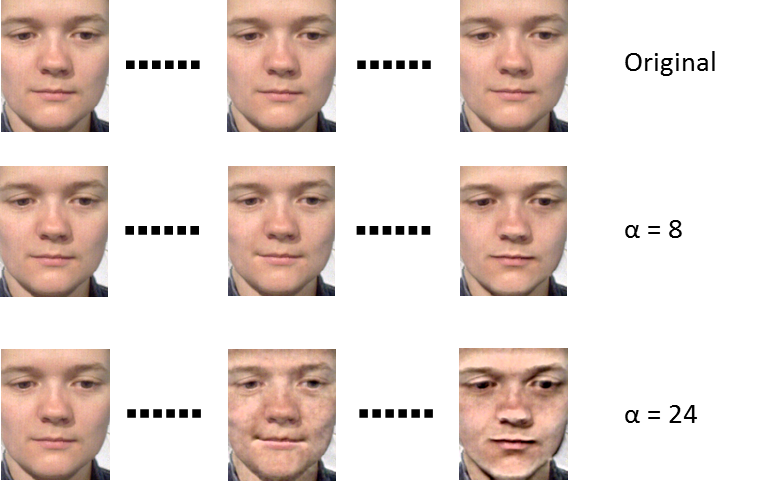}
\vspace{-2ex}
\caption{An ME clip magnified at different $\alpha$ levels.}
\label{fig:magnification}
\end{center}
\end{figure}

One major challenge for ME recognition is that the intensity levels of facial movements are too low to be distinguishable. 
To enhance the differences of MEs we propose to use the Eulerian video magnification method~\cite{wu2012eulerian} to magnify the subtle motions in videos. 
The original method was proposed for magnifying either motion or color content of a video. 
Here we apply it for motion magnification. 
{The magnification process is implemented using codes shared by authors of ~\cite{wu2012eulerian}.
$\alpha$ is a parameter that controls the level of motion amplification.
Bigger values of $\alpha$ lead to larger scale of motion amplification, but also can cause bigger displacement and artifacts. 
An example of ME clip magnified at different $\alpha$ levels is shown in Figure~\ref{fig:magnification}.
Effects of using different magnification factor $\alpha$ for ME recognition are explored in our experiment by varying $\alpha$ in ten levels.
For more details about the Eulerian video magnification method we refer readers to~\cite{wu2012eulerian}.}

\vspace{-2ex}

The Eulerian video magnification method facilitates ME recognition as the differences between different categories of MEs are enlarged. 
However, we do not use it for ME spotting because it magnifies unwanted motions (\eg head movements) at the same time. The issue is discussed in Section \uppercase\expandafter{\romannumeral5}.A.

\vspace{-2ex}
%%%%%%%%%%%%%%%%%%%%%
\subsection{Temporal interpolation model (TIM)}
Another difficulty for ME recognition is the short and varied duration. 
The problem is more evident when the videos are filmed with low frame rate. 
For example when recording at a standard speed of 25 fps, some MEs only last for four to five frames, which limit the application of some spatial-temporal descriptors, \eg, if we use LBP-TOP we can only use the radius $r=1$. 
To counter for this difficulty we propose to use the Temporal interpolation model (TIM) introduced by Zhou~\etal~\cite{zhou2011towards}. 

{The TIM method relies on a path graph to characterize the structure of a sequence of frames. A sequence-specific mapping is learned to connect frames in the sequence and a curve embedded in the path graph so that the sequence can be projected onto the latter. The curve, which is a continuous and deterministic function of a single variable $t$ in the range of [0,1], governs the temporal relations between the frames. Unseen frames occurring in the continuous process of an ME are also characterized by the curve. Therefore a sequence of frames after interpolation can be generated by controlling the variable $t$ at different time points accordingly.}

The TIM method allows us to interpolate images at arbitrary time positions using very small number of input frames. 
We use TIM to interpolate all ME sequences into the same length (\eg of 10, 20, ... or 80 frames) for two purposes: 
first is for up-sampling the clips with too few frames; 
secondly, with a unified clip length, more stable performance of feature descriptors can be expected. 
More details about the TIM method we refer readers to~\cite{zhou2011towards}. 
The effect of interpolation length is discussed in Section \uppercase\expandafter{\romannumeral5}.B.1.

\vspace{-2ex}
%%%%%%%%%%%%%%%%%%%%%
\subsection{Feature extraction}
Several spatial-temporal local texture descriptors have been demonstrated to be effective in tackling the FE recognition problem, especially LBP and its variants \cite{zhao2007dynamic,Dhall2013emotion,Kahou2014facial}. 
Here we employ three kinds of features in our ME recognition framework to compare their performance. 
Details of each descriptor are described below, and the comparison of their performance will be discussed in Section \uppercase\expandafter{\romannumeral5}.B.2.

\subsubsection{LBP on three orthogonal planes}
Local binary pattern on three orthogonal planes (LBP-TOP), proposed by Zhao and Pietik{\"a}inen~\cite{zhao2007dynamic}, is an extension of the original LBP for dynamic texture analysis in spatial-temporal domain. 
It is one of the most frequently used features for FE recognition, and also for recent ME recognition studies. 

\begin{figure}[H]
\begin{center}
\includegraphics[scale=0.5]{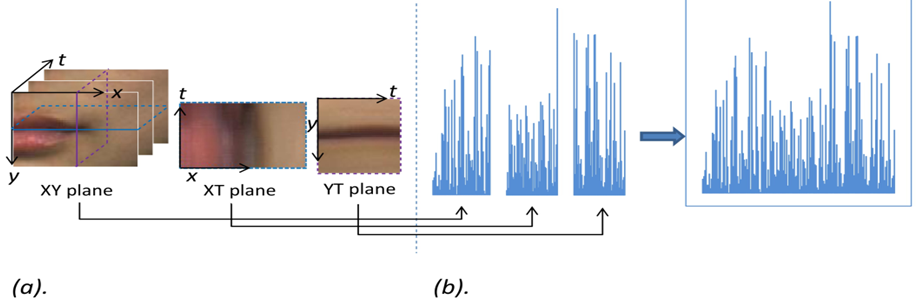}
\vspace{-2ex}
\caption{(a) The textures of XY, XT and YT planes, (b) Their corresponding histograms and the concatenated LBP-TOP feature.}
\label{fig:LBPTOP}
\vspace{-2ex}
\end{center}
\end{figure}

A video sequence can be thought as a stack of XY planes on T dimension, as well as a stack of XT planes on Y dimension, or a stack of YT planes on X dimension. 
The XT and YT plane textures can provide information about the dynamic process of the motion transitions. 
Figure~\ref{fig:LBPTOP}(a) presents the textures of XY, XT and YT plane around the mouth corner of one ME clip.

The traditional way of extracting LBP-TOP feature is illustrated in Figure~\ref{fig:LBPTOP}(b). 
To include information from 3D spatial-temporal domain, LBP codes are extracted from every pixel of XY, XT or YT plane to produce their corresponding histograms.
The three histograms are then concatenated into a single histogram as the final LBP-TOP feature vector. 

It can be seen that the histograms for XY, XT and YT plane represent different information, and previous results indicate that the traditional LBP-TOP feature which uses all three histograms does not always yield the best performance~\cite{davison2014micro}. 
In our method we will consider five different (combinations of) LBP histograms on the three planes as listed in Table~\ref{table:LBPcombinations}.

\begin{table}[H]
\caption{Five combinations of LBP features on three orthogonal planes and their corresponding abbreviations.}
\label{table:LBPcombinations}
\vspace{-4ex}
\begin{center}
\begin{tabular}{|l|l|}
\hline
\textbf{Abbreviation} & \textbf{Histogram of which plane}\\
\hline
LBP-TOP & XY + XT + YT\\
\hline
LBP-XYOT & XT + YT\\
\hline
LBP-XOT & XT\\
\hline
LBP-YOT & YT\\
\hline
LBP & XY\\
\hline
\end{tabular}
\end{center}
\end{table}
\vspace{-2ex}

\subsubsection{HOG and HIGO on three orthogonal planes}
Histograms of Oriented Gradients (HOG)~\cite{dalal2005histograms} is another popular feature descriptor which has been applied in FE recognition~\cite{deniz2011face,
li2009facial}. 
As reviewed in Section \uppercase\expandafter{\romannumeral2}.C, most previous ME recognition studies considered LBP feature. 
Here we use HOG as the second feature descriptor for ME recognition.

First, we formulate the 2D HOG on the XY plane. 
Provided an image $I$, we obtain the horizontal and vertical derivatives $I_x$  and $I_y$ using the convolution operation. More specifically $I_x=I*K^T$ and $I_y=I*K$, where $K\!=[-1\, 0\, 1]^T$. 
For each point of the image, its local gradient direction $\theta$ and gradient magnitude $m$ are computed as follows:
\begin{align}
\theta&=\arg(\nabla I)=\atan2(I_y,I_x), \\
m&=|\nabla I|=\sqrt{I_x^2+I_y^2}.
\end{align}

Let the quantization level for $\theta$ be $B$ and $\mathcal{B}=\{1,\dots,B\}$. 
Note that $\theta \in [-\pi, \pi]$. 
Thus a quantization function of $\theta$ is a mapping $Q:[-\pi,\pi]\to\mathcal{B}$. 
{As a result, for a local $2$D region (\ie a block) or a sequence of $2$D regions (\ie a cuboid) $\mathcal{N}$}, the histogram of oriented gradients (HOG)  is a function $g: \mathcal{B} \to R$. 
More specifically, it is defined by
\begin{equation}\label{eq:HOGfunction}
g(b)=\sum_{\bold{x}\in\mathcal{N}}\delta(Q(\theta(\bold{x})),b)\cdot m(\bold{x}),
\end{equation}
where $b\in \mathcal{B}$ and $\delta(i,j)$ is the Kronecker's delta function as
\vspace{-1ex}
\begin{equation}
\delta(i,j)= \left\{
 \begin{array}{rl}
  1 &  \text{if } i=j\\ 
  0 &  \text{if } i\neq j\,.
 \end{array} \right.
\end{equation}
For HOG, each pixel within the block or cuboid has a weighted vote for a quantized orientation channel $b$ according to the response found in the gradient computation.

Second, we introduce the third feature descriptor employed: the histogram of image gradient orientation (HIGO). 
HIGO is a degenerated variant of HOG: it uses `simple' vote rather than `weighted vote' when counting the responses of the histogram bins. In detail, the function $h$ for HIGO is defined as:
\vspace{-1ex}
\begin{equation}
h(b)=\sum_{\bold{x}\in\mathcal{N}}\delta(Q(\theta(\bold{x})),b),
\end{equation}
where $b$ and $\delta$ have the same meaning as in Equation~\eqref{eq:HOGfunction}.

{HIGO depresses the influence of illumination and contrast by ignoring the magnitude of the first order derivatives.  It is shown that the image gradient orientation $\theta(\bold{x})$ does not depend on the illuminance at pixel $\bold{x}$ \cite{zhang2009face}. Thus the histogram of image gradient orientation statistically obtained in an image block is illumination insensitive.}
For those challenging tasks (\eg recognizing spontaneous MEs recorded in authentic situations) in which the illumination conditions substantially vary, HIGO is expected to have enhanced performance.  

The corresponding 3D spatial-temporal version of HOG and HIGO features can be easily obtained by extending the descriptor on the XY plane to the three orthogonal planes, as is done when one extends LBP to LBP-TOP. 
We explore five different (combinations of) HOG and HIGO from the three orthogonal planes as we explained for the LBP feature. 
Their abbreviations are defined the same way as in Table~\ref{table:LBPcombinations} (with `LBP' changed to`HOG' or `HIGO').

In order to account for the variations in illumination or contrast, the gradient intensity is usually normalized. 
In this paper, we use the local L1 normalization for the histogram calculated from each block/cuboid. 
The concatenations of the normalized block/cuboid histograms are globally normalized (either L1 or L2) to form the final descriptor.

\vspace{-2ex}
%%%%%%%%%%%%%%%%%%%%%%%%%
\subsection{Classification}

We use a linear Support Vector Machine (LSVM)~\cite{chang2011libsvm} as the classifier for ME recognition. 
For the parameter selection of LSVM, we carried out a five-fold cross-validation on the training data for parameters search in the range of $[10^{-1}, 1,2, 10, \cdots, 10^3]$, and the parameter setting leading to the best performance was selected for that trail. 
For SMIC, ME samples are classified into three categories of positive, negative and surprise. 
For CASMEII, ME samples are classified into five categories of happiness, disgust, surprise, repression and others.

\vspace{-2ex}
%%%%%%%%%%%%%%%%%%%%%%%%%%%%%%%%%%%%%%%%%%%%%%%%%%%%%%%
\section{Experimental results and discussions}

We next present experiments and discuss our results.
Section \uppercase\expandafter{\romannumeral5}.A presents results for ME spotting; Section \uppercase\expandafter{\romannumeral5}.B explains ME recognition results in four sub-experiments; Section \uppercase\expandafter{\romannumeral5}.C tests a fully automatic ME analysis system which combines both spotting and recognition process; and Section \uppercase\expandafter{\romannumeral5}.D compares our automatic method to human subjects' performance on both ME spotting and ME recognition tasks.

\vspace{-3ex}
%%%%%%%%%%%%%%%%%%%%%%%%%
\subsection{ME spotting}

\subsubsection{Datasets}
SMIC and CASMEII databases are used in the spotting experiment. 
The original SMIC database only includes labeled ME clips that include frames from onset to offset. 
It is suitable for the ME recognition problem, but the spotting test involves longer video sequences which also include frames before and after the ME span. 
We re-built an extended version of SMIC (SMIC-E) by extracting longer clips around time points when MEs occur from the original videos. 
The three datasets in the new SMIC-E database with longer video sequences are denoted as SMIC-E-VIS, SMIC-E-NIR and SMIC-E-HS accordingly. 
The SMIC-E-VIS and SMIC-E-NIR dataset both include 71 long clips of average duration of 5.9 seconds. 
The SMIC-E-HS dataset contains 157 long clips of an average length of 5.9 seconds. 
{Four clips contain two MEs as they are located close to each other.}
Three samples from the original SMIC-HS dataset are not included in the SMIC-E-HS due to data loss of the original videos.
Unlike SMIC, the original CASMEII database provides long video clips that include extra frames before and after the ME span, so we are able to use the clips for ME spotting as they are. 
The average duration of CASMEII clips is 1.3 seconds.  
One clip in CASMEII was excluded because its duration is too short. 

\begin{figure*}[thbp]
\centering
\includegraphics[scale=0.6]{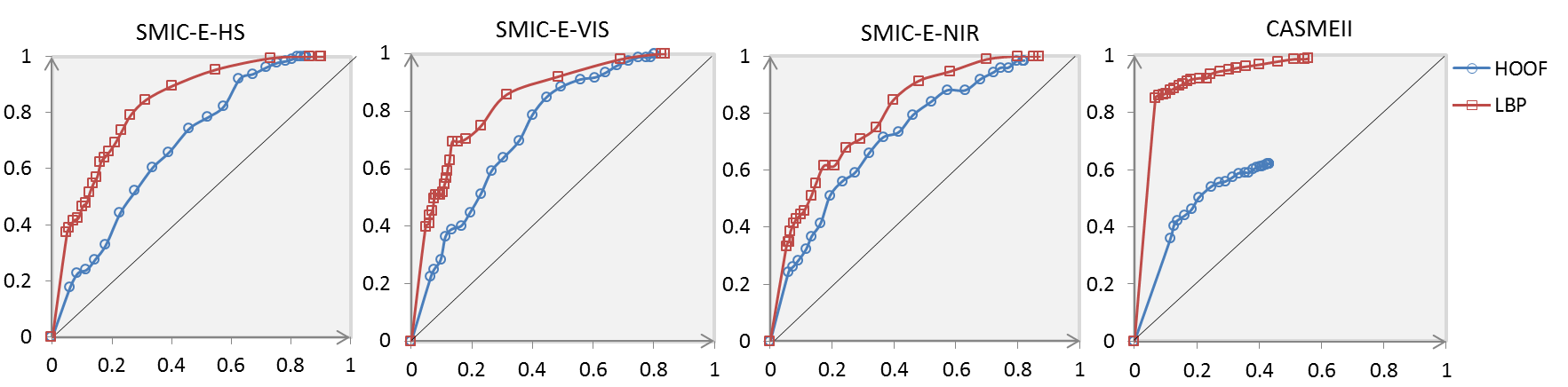}
\caption{ROC curves for ME spotting on CASMEII and three SMIC-E datasets. The $x$ axis shows the false positive rate (FPR), and the $y$ axis shows the true positive rate (TPR).}
\label{fig:result_spot_ROC}
\end{figure*}
\subsubsection{Parameters}

The ME interval $N$ is set to correspond to a time window of about 0.32 seconds ($N=9$ for SMIC-E-VIS and SMIC-E-NIR, $N=33$ for SMIC-E-HS, $N=65$ for CASMEII). 
For LBP, uniform mapping is used, the radius $r$ is set to $r=3$, and the number of neighboring points $p$ is set to $p=8$ for all datasets. 
For HOOF, the parameters were left to the default values as in~\cite{liu2009beyond}. 
We tested two different approaches for selecting the reference frame. 
In the first approach, the reference frame was fixed to be the first frame of the video.
In the second approach, the reference frame was set to be the TF, \ie the reference frame follows the CF in a temporal sliding window manner. 
Results showed that the first approach always yields better performance than the second one -- therefore in the following we report results using the first approach. 
We also tried using motion magnification in ME spotting in another prior test. 
We magnified videos before extracting features. 
However, it turned out that both true and false peaks were enlarged and the spotting accuracy was not improved. 
Therefore motion magnification is not used in the following ME spotting experiments.

After the peak detection all the spotted peak frames are compared with ground truth labels to tell whether they are true or false positive spots. 
With a certain threshold level, if one spotted peak is located within the frame range of $[\text{onset}-(N-1)/4, \text{offset}+(N-1)/4]$ of a labeled ME clip, the spotted sequence will be considered as one true positive ME; otherwise the $N$ frames of spotted sequence will be counted as false positive frames. 
We define the true positive rate (TPR) as the percentage of frames of correctly spotted MEs, divided by the total number of ground truth ME frames in the dataset.
The false positive rate (FPR) is calculated as a percentage of incorrectly spotted frames, divided by the total number of non-ME frames from all the long clips.  
Performance of ME spotting is evaluated using receiver operating characteristic (ROC) curves with TPR as the $y$ axis and the FPR as the $x$ axis.

\subsubsection{Results}
The proposed method was tested on CASMEII and the three datasets of SMIC-E. 
The spotting results on each dataset are presented in Figure~\ref{fig:result_spot_ROC}. 
A ROC curve is drawn for each of the two feature descriptors on one dataset. 
Points of the ROC curves in Figure~\ref{fig:result_spot_ROC} are drawn by varying the percentage parameter $\tau$ (in equation \ref{eq:spot_T}) from 0 to 1 with step size of 0.05. 

From the figure, we observe that more MEs are correctly spotted when we drop the threshold value, but with the expense of higher FPR. 
The area under the ROC curve (AUC) for each curve is calculated and listed in Table~\ref{table:result_spot_AUC}. 
Higher AUC value indicates better spotting performance. 

\begin{table}[H]
\caption{AUC values of the ME spotting experiments using LBP and HOOF as feature descriptors on CASMEII and three datasets of SMIC-E.}
\label{table:result_spot_AUC}
\vspace{-2ex}
\centering
\begin{tabular}{|l|c|c|c|c|}
\hline
  & SMIC-E-HS & SMIC-E-VIS & SMIC-E-NIR &  CASMEII \\
\hline
LBP & 83.32\% & 84.53\% & 80.60\% & 92.98\% \\
\hline
HOOF & 69.41\% & 74.90\% & {73.23\%} & 64.99\% \\
\hline
\end{tabular}
\end{table}

Figure~\ref{fig:result_spot_ROC} and Table~\ref{table:result_spot_AUC} show that LBP outperforms HOOF for the proposed ME spotting method, as its AUC values are higher and the FPRs are lower. 
For spotting on the three datasets of SMIC-E, best performance is achieved on SMIC-E-VIS dataset. 
By using LBP, our proposed method can spot about 70\% of MEs with only 13.5\% FPR, and the AUC is 84.53\%. 
We observe that the majority of the false positives are eye blinks (discussed in detail below).
On CASMEII, the advantage of LBP feature is more obvious (AUC of 92.98\%). 
We hypothesize that the reason why a higher AUC is achieved on CASMEII is that CASMEII contains shorter video clips than SMIC-E (so the spotting task is easier).

This is the first report of ME spotting result on spontaneous ME databases SMIC and CASMEII, so there are no results to compare with. 
The current results show that spontaneous MEs can be spotted by comparing the feature differences of the CF and the AFF, and LBP is more efficient than HOOF. 
Spotting MEs in spontaneous videos is significantly more difficult than on posed videos, as random motions could interfere as noise. 

\begin{figure}[H]
\begin{center}
\includegraphics[scale=0.6]{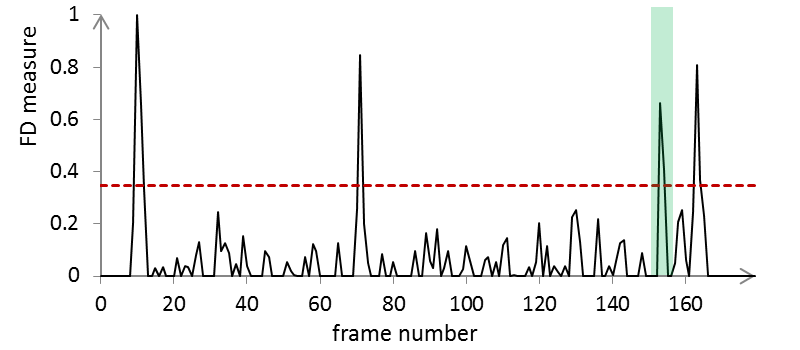}
\vspace{-2ex}
\caption{An example of ME spotting result showing a true spot of ME (the green peak), with three false spots of eye blinks.}
\label{fig:eyeblinks}
\end{center}
\end{figure}
\vspace{-2ex}

\paragraph*{{\bf Discussion of failure cases}}
Upon detailed examinations of the spotting result, we found that a large portion of false spots were caused by eye movements, such like eye blinks or eye-gaze changes.
One such example is shown in Figure~\ref{fig:eyeblinks} (the spotting result on a video from SMIC-E-VIS). 
Four peaks were spotted when the threshold level was set to the dashed line. 
We compared the peaks to the ground truth labels and found that only the third peak (around frame 153) is a true positive spot of ME, while the other three peaks were caused by eye blinks. 
As we report results on spontaneous facial videos, it is expected that there will be many eye blinks in the videos. 
The proposed method detects transient movements on the scale of the whole face, so eye movements could also be detected as big peaks. 
In contrast, in posed databases eye blinks are significantly less common as they can be voluntarily concealed.
Therefore this issue has never been addressed in previous ME spotting works.

We tried two ways to rule out eye movements:
(i)~by excluding the eye regions during feature extraction; and 
(ii)~by using an eye-blink-detector to exclude eye blink peaks from the results.
Both approaches helped to reduce the FPRs. 
However, both approaches also caused a decrease in TPR at the same time. 
This is due to many spontaneous MEs involving muscles around eye regions.
Furthermore, the onsets of many MEs (about 50\%, as we have empirically measured) also temporally overlap with eye blinks.
Thus neither approach turned out to be good enough. 
We plan to perform more comprehensive investigations about this issue in future work. 

\vspace{-1ex}

%%%%%%%%%%%%%%%%%%%%%%%%%
\subsection{ME recognition}
In this section we report results of ME recognition experiments on the full version of SMIC and CASMEII. 
ME clips (including frames from onset to offset) with raw images are used. 
We use leave-one-subject-out protocol in all of our ME recognition experiments.
We therefore carry out four sub-experiments to evaluate the effect of each of these factors:
(1) the effect of temporal interpolation; (2) different feature descriptors; (3) different modalities in the datasets; and (4) the effect of the motion magnification method. Finally, we provide a comparison to the state of the art.

%%%%%%%%%%%%%%%%%%%%%%%%%%
\subsubsection{Effect of the interpolation length}
The aim of this first sub-experiment is to explore how the sequence length of interpolation affects the accuracy of ME recognition. 

\paragraph*{{\bf Parameters}}
In order to control the effect from other factors and focus on TIM, we skip the motion magnification step, and use only LBP-TOP as the feature descriptor (with a group of fixed parameters).
After face alignment, TIM is applied to interpolate ME sequences into eight different lengths (10, 20, 30, \dots, 80), and then LBP-TOP features ($8\time8\times1$ blocks, $r=2$, $p=8$) are extracted. 
We test on SMIC-HS, SMIC-VIS and SMIC-NIR datasets. 
The average sequence length of the original ME samples is 33.7 frames for SMIC-HS and 9.66 frames for SMIC-VIS and SMIC-NIR. 

\begin{figure}[H]
\begin{center}
\includegraphics[scale=0.6]{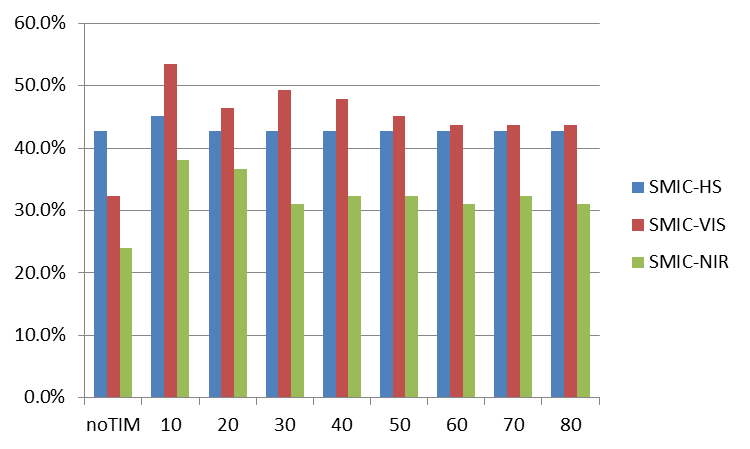}
\vspace{-2ex}
\caption{ME recognition accuracy with different TIM lengths, using LBP-TOP as feature descriptor. The $x$-axis shows the frame numbers of ME sequences after TIM interpolation, and the $y$-axis shows accuracies.}
\label{fig:result_rec_TIM}
\end{center}
\end{figure}

\paragraph*{{\bf Results}}
Figure~\ref{fig:result_rec_TIM} shows results on the three datasets at different sequence lengths. 
The best performance for all three datasets is achieved with interpolation to 10 frames (TIM10). 
We analyzed this finding from two aspects:

First, the left side of the figure shows that TIM10 leads to higher performance than no TIM (without interpolation). 
For the VIS and NIR datasets, TIM10 hardly alters the sequence lengths, as the average length of the no TIM sequences (original ME samples) is 9.66 frames. 
For the HS dataset, TIM10 is a down-sampling process, as the average length of original sequences is 33.7 frames. 
We therefore hypothesize that TIM10 performs better than no TIM because the input frames with unified sequence length improve the performance of the feature descriptor. 

Secondly, the right side of the figure shows that longer interpolated sequences (TIM20 to TIM80) do not lead to improved performance compared to TIM10. 
With the current datasets using TIM method, it appears that interpolation to 10 frames is enough. 
A possible explanation for the performance here is that in longer sequences, changes along the time-dimension are diluted, which makes feature extraction more challenging. 
This finding is supported by our previous reported in~\cite{pfister2011recognising} and~\cite{li2013spontaneous}.   

\begin{table}[H]
\caption{Computation time comparison for different TIM lengths.}
\vspace{-2ex}
\label{table:TIMrunningtime}
\begin{center}\tabcolsep=0.08cm
\begin{tabular}{|c|c|c|c|c|c|c|c|c|c|}
\hline
TIM length & noTIM & 10 & 20 & 30 & 40 & 50 & 60 & 70 & 80\\
\hline
Time (s) & 87.8 & 22.1 & 49.6 & 77.2 & 104.9 & 132.9 & 160.0& 188.3 & 216.6\\
\hline
\end{tabular}
\end{center}
\end{table}

We also evaluate the running time for different TIM lengths.
These results are shown in Table~\ref{table:TIMrunningtime}. 
The higher interpolation frame rate requires higher computation time (and also occupies more storage space, if the frames are stored) but does not consistently lead to improved accuracy. 

To conclude, we see evidence that TIM process is necessary and valuable for ME recognition, and that 10 frames is enough. 
We therefore set TIM length as 10 for all the following experiments.

%%%%%%%%%%%%%%%%%%%%%%%%%%
\subsubsection{Comparison of three features}

In this sub-experiment, we evaluate the performance of three features (LBP, HOG and HIGO) for the ME recognition task. 
Multiple combinations of histograms on three orthogonal planes are considered for the three features respectively (see Table~\ref{table:LBPcombinations}).

\paragraph*{{\bf Parameters}}
After face alignment, the sequences are interpolated into 10 frames using TIM method (the step of motion magnification is skipped in this experiment for later discussion in sub-experiment 4). 
For feature extraction, three kinds of features are extracted from evenly divided blocks of each interpolated sequence. 
Multiple groups of features are extracted by varying several parameters. 
For the LBP feature, we vary the radius $r$, neighbor points $p$ and the number of divided blocks; for the HOG and HIGO features, we vary the number of divided blocks, and the number of bins is fixed as $b=8$. 
Training and testing is done using linear SVM with leave-one-subject-out protocol. 
Experiments are conducted on CASMEII and the three datasets of SMIC.
\begin{table*}[t]
\caption{ME recognition results using LBP, HOG and HIGO as feature descriptors on each dataset. LSVM is employed as the classifier using leave-one-subject-out protocol. The highest accuracies for each feature on each dataset are marked in bold font. $(p,r)$ indicates the neighbor points $p$ and radius $r$ of LBP feature; $b$ is the number of bins of HIGO and HOG features.}
\label{table:reg_3features}
\tabcolsep=0.08cm
\centering
\begin{tabular}{|c|c|c|c|c|c|c|c|c|c|c|c|c|c||c|c|c|}
\hline
\multicolumn{2}{|c|}{} & \multicolumn{3}{c|}{CASMEII} & \multicolumn{3}{c|}{SMIC-HS} & \multicolumn{3}{c|}{SMIC-VIS} & \multicolumn{3}{c||}{SMIC-NIR} &\multicolumn{3}{c|}{SMIC-subHS}\\
\cline{3-17}
\multicolumn{2}{|c|}{} & \multirow{2}{*}{block}  & Acc.  & $(p,r)$ & \multirow{2}{*}{block} & Acc.  & $(p,r)$  & \multirow{2}{*}{block} & Acc.  & $(p,r)$  & \multirow{2}{*}{block} & Acc.  & $(p,r)$ & \multirow{2}{*}{block} & Acc.  & $(p,r)$ \\
\multicolumn{2}{|c|}{} &   &  (\%) & or $b$ &   & (\%) & or $b$ &   & (\%) & or $b$ &   & (\%) & or $b$ &   &  (\%) & or $b$\\
\hline\hline

\multirow{5}{*}{LBP-}
 & TOP & 8$\times$8$\times$2 & \textbf{55.87} &  (8,2) & 8$\times$8$\times$2 & 51.83 &  (8,2) & 5$\times$5$\times$1 & \textbf{70.42} &  (8,2) & 5$\times$5$\times$1 & \textbf{64.79} &  (8,3) & 8$\times$8$\times$2 & 76.06 &  (8,2)\\
 & XYOT & 8$\times$8$\times$4 & \textbf{55.87} &  (8,2) & 8$\times$8$\times$2 & 56.10 &  (8,1) & 5$\times$5$\times$1 & \textbf{70.42} &  (8,2) & 8$\times$8$\times$2 & \textbf{64.79} &  (8,2) & 8$\times$8$\times$2 & \textbf{77.46} &  (8,2)\\
 & XOT & 8$\times$8$\times$4 & 55.06 &  (8,2) & 8$\times$8$\times$2 & \textbf{57.93} &  (8,1) & 5$\times$5$\times$1 & \textbf{70.42} &  (8,2) & 8$\times$8$\times$1 & 54.93 &  (8,2) & 5$\times$5$\times$2 & \textbf{77.46} &  (8,2)\\
 & YOT & 5$\times$5$\times$4 & 54.85 &  (8,1) & 8$\times$8$\times$2 & 50.61 &  (8,1) & 5$\times$5$\times$2 & \textbf{70.42} &  (8,1) & 8$\times$8$\times$4 & \textbf{64.79} &  (8,2) & 8$\times$8$\times$2 & 76.06 &  (8,2)\\
 & LBP & 8$\times$8$\times$2 & 44.53 &  (8,2) & 8$\times$8$\times$2 & 43.29 &  (8,2) & 5$\times$5$\times$1 & 67.61 &  (8,2) & 8$\times$8$\times$4 & 50.70 &  (8,2) & 8$\times$8$\times$2 & 64.69 &  (8,2)\\

\hline
\multirow{5}{*}{HIGO-} 
 & TOP & 8$\times$8$\times$2 & 55.87 &  8 & 6$\times$6$\times$2 & 59.15 &  8 & 4$\times$4$\times$2 & 69.01 &  8 & 6$\times$6$\times$2 & 53.52 &  8 & 4$\times$4$\times$2 & \textbf{80.28} &  8\\
 & XYOT & 8$\times$8$\times$2 & 55.47 &  8 & 6$\times$6$\times$2 & 59.76 &  8 & 6$\times$6$\times$2 & 71.83 &  8 & 6$\times$6$\times$1 & 52.11 &  8 & 4$\times$4$\times$2 & 78.87 &  8\\
 & XOT & 8$\times$8$\times$2 & 53.44 &  8 & 4$\times$4$\times$2 & \textbf{65.24} &  8 & 6$\times$6$\times$2 & \textbf{76.06} &  8 & 6$\times$6$\times$1 & 47.89 &  8 & 4$\times$4$\times$2 & 78.87 &  8\\
 & YOT & 8$\times$8$\times$2 & \textbf{57.09} &  8 & 6$\times$6$\times$2 & 58.54 &  8 & 4$\times$4$\times$2 & 71.83 &  8 & 6$\times$6$\times$2 & \textbf{59.15} &  8 & 4$\times$4$\times$2 & 78.87 &  8\\
 & HIGO & 8$\times$8$\times$2 & 42.51 &  8 & 2$\times$2$\times$8 & 50.61 &  8 & 4$\times$4$\times$1 & 60.56 &  8 & 6$\times$6$\times$2 & 35.21 &  8 & 4$\times$4$\times$1 & 64.79 &  8\\

\hline
\multirow{5}{*}{HOG-} 
 & TOP & 8$\times$8$\times$2 & \textbf{57.49} &  8 & 2$\times$2$\times$2 & \textbf{57.93} &  8 & 2$\times$2$\times$2 & 67.61 &  8 & 2$\times$2$\times$8 & \textbf{63.38} &  8 & 4$\times$4$\times$2 & \textbf{80.28} &  8\\
 & XYOT & 8$\times$8$\times$2 & \textbf{57.49} &  8 & 2$\times$2$\times$2 & 51.83 &  8 & 6$\times$6$\times$2 & \textbf{71.83} &  8 & 2$\times$2$\times$2 & 60.56 &  8 & 6$\times$6$\times$6 & 71.83 &  8\\
 & XOT & 8$\times$8$\times$2 & 51.01 &  8 & 4$\times$4$\times$8 & \textbf{57.93} &  8 & 4$\times$4$\times$2 & \textbf{71.83} &  8 & 6$\times$6$\times$2 & 56.34 &  8 & 2$\times$2$\times$2 & 69.01 &  8\\
 & YOT & 8$\times$8$\times$2 & 56.68 &  8 & 2$\times$2$\times$2 & 51.22 &  8 & 6$\times$6$\times$2 & 67.61 &  8 & 2$\times$2$\times$8 & 59.15 &  8 & 6$\times$6$\times$2 & 69.01 &  8\\
 & HIGO & 8$\times$8$\times$2 & 40.49 &  8 & 2$\times$2$\times$2 & 52.44 &  8 & 6$\times$6$\times$2 & 54.93 &  8 & 2$\times$2$\times$2 & 53.52 &  8 & 6$\times$6$\times$2 & 60.56 &  8\\
\hline
\end{tabular}
\end{table*}
\setlength{\textfloatsep}{10pt}

\paragraph*{{\bf Results}}
The ME recognition results using three feature descriptors are shown in Table~\ref{table:reg_3features} for each dataset. 
For each feature, we list results of using five combinations from three orthogonal planes. 
For each combination, the best result with its corresponding parameters are listed. 
We discuss the first four columns of results in this section.
The last column of results for SMIC-subHS dataset will be discussed in sub-experiment 3. 

Two conclusions can be drawn from the results of the first four columns of Table~\ref{table:reg_3features}.
First, TOP (three orthogonal planes) does not consistently yield the best performance for ME recognition. 
Sometimes better results are achieved from using only XOT, YOT or XYOT. 
This finding is consistent for all three features on all four test datasets. 
This indicates that the dynamic texture along the T dimension represents the most important information for ME recognition. 
On the other hand, the XY histogram seems to contain redundant information about the facial appearance rather than motion, thus making classification more difficult. 
Similar findings were also reported in~\cite{davison2014micro}. 
Secondly, the gradient-based features HOG and HIGO outperform LBP for ME recognition on ordinary RGB data (CASMEII, SMIC-HS and SMIC-VIS). 
The best result obtained on SMIC is 76.06\% using HIGO-XOT. 
Further comparison between the two gradient based features shows that HIGO performs better than HOG. 
One possible explanation for this is that HIGO is invariant to the magnitude of the local gradients, which varies significantly across subjects due to different muscle moving speeds.
However, for infrared data (SMIC-NIR) this trend is different: LBP performs best. 
The textures recorded by an NIR camera are very different from RGB videos, as NIR textures are less affected by illumination. 
This is consistent with another study~\cite{zhao2011facial} which also reported that LBP feature is suitable for NIR data .

%%%%%%%%%%%%%%%%%%%%%%%%%%
\subsubsection{Comparison of datasets recorded with different cameras}

In this sub-experiment, we compare the ME recognition performance using different recording instruments (the SMIC dataset includes ME samples recorded with three cameras). 
In sub-experiment 2, the SMIC-VIS dataset led to best performance on SMIC.
However, SMIC-HS contains more ME samples than the SMIC-VIS and SMIC-NIR. 
To make the comparison fair, we form a SMIC-subHS dataset containing the same 71 ME samples as SMIC-VIS and SMIC-NIR, and run the same test as we did in sub-experiment 2.

\paragraph*{{\bf Results}}
The results are shown in the rightmost column in Figure~\ref{table:reg_3features}. 
By comparing the results of SMIC-VIS, SMIC-NIR and SMIC-subHS which contain the same samples we found: 

First, SMIC-subHS dataset yields the highest accuracy of 80.28\%.
This demonstrates that the worse performance on the SMIC-HS dataset was due to it including more (possibly distracting) samples. 
By comparing results of SMIC-subHS and SMIC-VIS, we observe that camera recording at higher frame rate does facilitate the ME recognition as claimed in~\cite{yan2014casme}. 
So using a high speed camera is beneficial for automatically analyzing ME. 

Secondly, ME recognition performance on the SMIC-NIR dataset is the lowest. 
Compared to faces in RGB videos, faces recorded by an NIR camera  lack skin textures. 
Our initial motivation for adding an NIR camera in the recording of the SMIC dataset was because NIR camera is less affected by illumination and shadows. 
Comparing the three kinds of features, LBP works better with NIR data than the other two features. 
But the overall performance achieved on the NIR dataset is unsatisfactory. 
Other methods need to be explored in the future for ME recognition on NIR data.

%%%%%%%%%%%%%%%%%%%%%%%%%%
\subsubsection{Motion magnification}
In the above sub-experiments we skipped the motion magnification step of our method to untangle the effect of different components. 
In this sub-experiment, we focus on motion magnification.
We show that Eulerian motion magnification can improve ME recognition performance.

\begin{figure*}[thbp]
\centering
\includegraphics[scale=0.55]{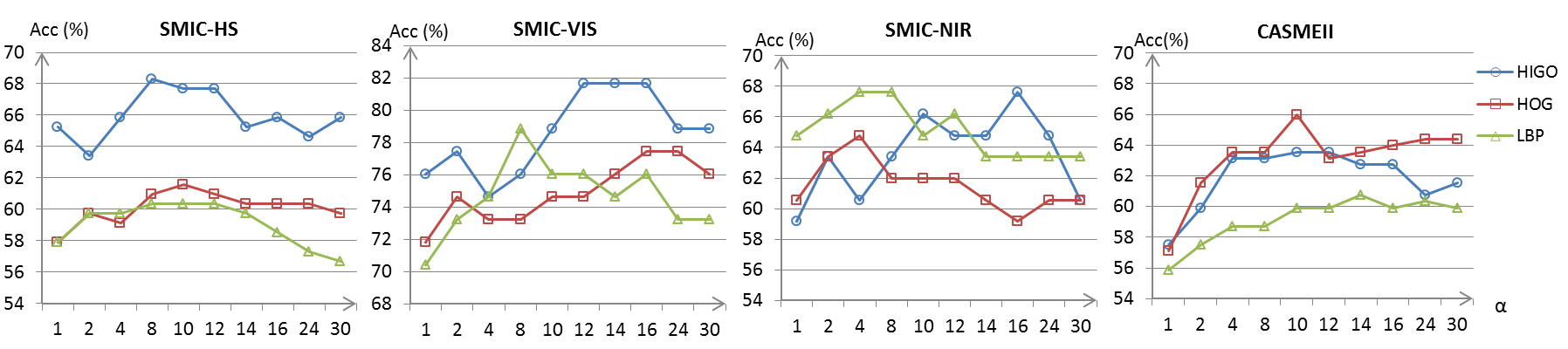}
\vspace{-1ex}
\caption{ME recognition results on SMIC and CASMEII databases at ten different motion magnification levels. The $x$-axis shows level of the magnification factor $\alpha$ ($\alpha=1$ indicates no magnification), and the $y$-axis shows the recognition accuracy.}
\label{fig:result_rec_magnification}
\end{figure*}

\paragraph*{{\bf Parameters}}
For this sub-experiment we apply all steps in the diagram of Figure~\ref{fig:rec_diagram}. 
After face alignment, we magnify the ME clips at ten different levels with $\alpha$ = 1, 2, 4, 8, 12, 16, 20, 24 and 30. 
Then the magnified clips are interpolated with TIM10, and the feature extraction and classification procedure are the same as we did in sub-experiments 2 and 3. 
The sub-experiment is carried on SMIC and CASMEII databases using LBP, HOG and HIGO features.

\paragraph*{{\bf Results}}
The results are shown in Figure \ref{fig:result_rec_magnification}.
One curve is drawn for each feature on each dataset.
The best performance achieved at each magnification level is presented in the curves. 
The figure shows that, compared to the baseline with no magnification ($\alpha = 1$), ME recognition performance is generally improved when motion magnification is applied.
This finding is consistent for all three kinds of features on all four testing datasets. 
The level of improvement fluctuates with the change of $\alpha$ value. 
The curves are rainbow-shaped, and the best performance is generally achieved when the motion is magnified in the range of $[8, 16]$. 
Magnification at lower levels might not be enough to reveal the ME motion progress; on the other hand, magnification at higher levels degrades the performance because too many artifacts are induced as shown in Figure \ref{fig:magnification}.

The final results of each feature on each dataset are summarized in Table~\ref{table:reg_magnification}. 
We observe that:
First, motion magnification substantially increases ME recognition performance; 
Secondly, results with motion magnification support our former findings in Experiment 2 that HIGO feature outperforms HOG and LBP for color camera recorded datasets, while for the NIR dataset, HIGO and LBP work equally well. 

\begin{table}[thbp]
\caption{ME recognition results (with and without magnification) of our method compared to state-of-the-art.}
\label{table:reg_magnification}
\vspace{-3ex}
\centering
\begin{tabular}{|l|c|c|c|c|}
\hline
  & SMIC-HS & SMIC-VIS & SMIC-NIR &  CASMEII \\
\hline\hline
LBP & 57.93\% & 70.42\% & 64.79\% & 55.87\% \\
\hline
LBP+Mag & 60.37\% & 78.87\% & \textbf{67.61\%} & 60.73\% \\
\hline
HOG & 57.93\% & 71.83\% & 63.38\% & 57.49\% \\
\hline
HOG+Mag & 61.59\% & 77.46\% & 64.79\% & 63.97\% \\
\hline
HIGO & 65.24\% & 76.06\% & 59.15\% & 57.09\% \\
\hline
HIGO+Mag & \textbf{68.29\%} & \textbf{81.69\%} & \textbf{67.61\%} & \textbf{67.21\%} \\
\hline
HIGO+Mag* & \textbf{75.00\%}* & \textbf{83.10\%}* & \textbf{71.83\%}* & \textbf{78.14\%}* \\
\hline\hline
Li \cite{li2013spontaneous} & 48.8\% & 52.1\% & 38.0\% & N/A \\
\hline
Yan \cite{yan2014casme} & N/A & N/A & N/A & 63.41\%* \\
\hline
Wang \cite{wang2014micro} & 71.34\%* & N/A & N/A & 65.45\%* \\
\hline
Wang \cite{wang2014LBP} & 64.02\%* & N/A & N/A & 67.21\%* \\
\hline
Wang~\cite{wang2015micro} & N/A & N/A & N/A & 62.3\%\\
\hline
Liong \cite{liong2014optical} & 53.56\% & N/A & N/A & N/A \\
\hline
Liong \cite{liong2014subtle} & 50.00\% & N/A & N/A & 66.40\%* \\
\hline
\multicolumn{5}{l}{\textit{* results achieved using leave-one-sample-out cross validation.}}
\end{tabular}
\end{table}
\setlength{\textfloatsep}{8pt}

\subsubsection{Comparison to the state of the art}
The best performance of our method is with motion magnification, TIM10 and HIGO features. 
By combining these three components, we achieved 81.7\% accuracy for ME recognition on the SMIC-VIS dataset, 68.29\% on the SMIC-HS dataset, 67.61\% on SMIC-NIR dataset, and 67.21\% on CASMEII. 
We list state-of-the-art results for these datasets in the table for comparison. 
We performed all experiments using the leave-one-subject-out validation protocol, while some of the reference results (for CASMEII, all reference results) were achieved using leave-one-sample-out validation protocol, which is much easier. 
For direct comparison with those results, we also added one row of results using \emph{HIGO+Mag with leave-one-sample-out protocol}. 

Previous work generally use SMIC-HS and CASMEII, while the lower frame rate versions of SMIC (SMIC-VIS and SMIC-NIR) are less explored. 
For SMIC-VIS and SMIC-NIR datasets, compared to baseline results reported in~\cite{li2013spontaneous}, we achieve an improvement of almost 30\%. 
For SMIC-HS and CASMEII, more reference results are listed, and our results are consistently better regardless of the evaluation protocols.
Based on these comparisons, our proposed framework outperforms all previous methods on all the four ME datasets.

\vspace{-2ex}
%%%%%%%%%%%%%%%%%%%%%
\subsection{An automatic ME analysis system (MESR) combining Spotting and Recognition }

\begin{figure}[thbp]
\begin{center}
\includegraphics[scale=0.53]{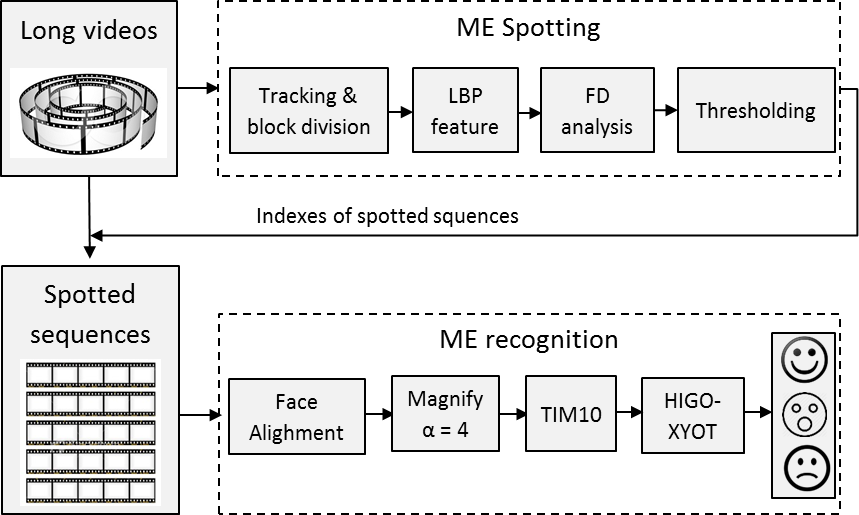}
\vspace{-1ex}
\caption{Framework of our automatic ME analysis system (MESR).}
\label{fig:MESR_diagram}
\end{center}
\end{figure}

Previous work has focused purely on either ME spotting or ME recognition, always considering these two tasks separately.
However, in reality, these two tasks have to be \emph{combined} to detect MEs in arbitrary long videos. 
We propose a complete ME analysis system (MESR) which first spots MEs in long videos, {and then classifies the spotted MEs into three categories of positive, negative and surprise. }
The flow of the proposed MESR method is shown in Figure~\ref{fig:MESR_diagram}. 
This MESR system works subject-independently (each input video is treated as an `unknown' test sample, and the classifier is trained on labeled MEs of the other subjects).

\setlength{\textfloatsep}{2pt}

\paragraph*{{\bf Parameters}}
Given a long video clip as the input for ME analysis, the MESR system first finds the locations at which an ME might occur by following the four steps of our ME spotting method described in Section \uppercase\expandafter{\romannumeral3}. LBP is selected for FD analysis.
The indexes of spotted frames are fed back to the original videos to excerpt short sequences for ME recognition. 
In the recognition process of MESR system, we use raw data and perform face alignment (instead of directly using spotted sequences) to register faces to the same model (which makes classification of the MEs easier). 
According to previous findings, we set the magnification level as $\alpha=4$, interpolation length as TIM10, use HIGO-XYOT as the feature descriptor, and linear SVM as the classifier for 3-classes classification. 

\paragraph*{{\bf Results}}
We test the MESR system on the SMIC-E-VIS dataset. 
The output of the spotting process can be varied by adjusting the threshold value, and a higher true spot rate is consistently associated with a higher false spot rate. 
In our experiments, we select the result at TPR $= 74.86\%$ (corresponding FPR $= 22.98\%$, $\tau = 0.15$) and all spotted sequences are all fed into the ME recognition component. 
For the correctly spotted ME sequences, the ME recognition component achieves $56.67\%$ accuracy for emotion recognition. 
The recognition accuracy drops comparing to the results in Experiment 4, which is expected as in the previous experiments all the MEs are hand-labeled, and thus the onset and offset time points are accurate. 
In contrast, in the MESR system we use automatically spotted sequences which do not always locate the MEs precisely, and include some non-ME frames. 
The overall performance of the MESR system is a multiplication of the two, \ie $Acc_{MESR} = 74.86\%\times56.67\% = 42.42\%$. 

We next compare the performance of our automatic ME analysis system to the performance of human subjects, and show that our automatic method performs comparatively to humans at this difficult task.
\vspace{-2ex}
%%%%%%%%%%%%%%%%%%%%%
\subsection{Human test results}
ME spotting and recognition is very challenging for ordinary people. 
In this section we experiment on human subjects on two tasks, and compare the results to our automatic method.

\subsubsection{Human test of ME recognition}
The first experiment concerns ME recognition. 
15 subjects (average age 28.5 years) were enrolled in the experiments (ten males, five female). 
All subjects signed consents which allow their data to be used for academic research and publications. 
71 ME samples from the SMIC-VIS dataset were used as the experiment videos. 
Before test started, the definition of an ME was explained, and three sample videos, each containing one class of ME, were shown to the subject. 
During the experiment, ME clips were shown on a computer screen one by one, and after each clip subjects selected which emotion category (positive, negative or surprise) the ME belongs to by clicking the corresponding button. 
The mean accuracy of the 15 human subjects was $72.11\% \pm 7.22\%$. 
In contrast, the best accuracy of our ME recognition method is 81.69\%.
This shows that our proposed ME recognition method outperforms human subjects in the ME recognition task.

\subsubsection{Human test of ME spotting \& recognition}
In the second experiment, we asked subjects to first spot whether there are any MEs in a video, and then indicate their emotion category (if there are any). 
Another 15 subjects (average age is 25.8 years) were enrolled (12 male, three female). 
All subjects signed consents which allow using their data for academic research and publications. 
71 long clips from the SMIC-E-VIS dataset, plus five neutral clips which do not include any MEs, were used as the experiment videos. 
Before test started, the definition of an ME was explained, and three example videos, each containing one class of ME, were shown to the subject. During the test, clips were shown on a computer screen one by one.
The subject first reported how many MEs (s)he spotted from the clip by clicking a button (either `0', `1', `2' or `more than 2'). 
If a button other than `0' was clicked, the subject was asked to further select which emotion category (positive, negative and surprise) the spotted ME(s) show. 
The subjects were informed that multiple MEs occurring near each other present the same emotion, so only one emotion category could be selected (even multiple MEs were spotted) for one clip.

We calculate an accuracy and a FPR for each subject, and then compute means and standard deviations for the group. 
For each subject, we count the number of MEs that were correctly spotted and recognized. 
The accuracy is a percentage of that number divided by 71. 
FPR is calculated as a percentage of false spots divided by the maximum possible false spots (152). 
The result shows that ME recognition ability varies across individuals.
The mean accuracy of the human subjects is $49.74\% \pm 8.04\%$ and the mean FPR is $7.31\% \pm 5.45\%$. 

In previous subsection our automatic MESR system achieved an accuracy of 42.42\%, which is a comparable performance (within one standard deviation) to the mean accuracy of the human subjects. 
Our method's main shortage is the false spot rate ($22.98\%$), which could be significantly reduced \eg if it can exclude non-ME fast movements such as eye blinks. 
As the first fully automatic system for ME analysis our method is a very promising start; improvement can be made in the future.

\vspace{-2ex}
%%%%%%%%%%%%%%%%%%%%%%%%%%%%%%%%%%%%%%%%%%%%%%%%%%%%%%%
\section{Conclusions}

In this work we focused on the study of \emph{spontaneous} MEs, which are much more difficult to analyze than posed expressions explored in previous work. 
We proposed novel methods for both ME \emph{spotting} and ME \emph{recognition}. 
For ME spotting, we are the first to propose a method able to spot MEs from spontaneous long videos. 
The method is based on feature difference (FD) comparison.
Two features (LBP and HOOF) are employed, and LBP is shown to outperform HOOF on two databases. 
For ME recognition, we proposed a new framework where motion magnification is employed to counter the low intensity of MEs. 
We validated the new framework on SMIC and CASMEII databases, and showed that our method outperforms the state of the art on both databases.
We also drew many interesting conclusions about the respective benefits of our method's components.
Finally, we proposed the first automatic ME analysis system which first spots and then recognizes MEs. 
It outperforms humans at ME recognition by a significant margin, and performs comparably to humans at the combined ME spotting \& recognition task.
This method has many potential applications such as in lie detection, law enforcement and psychotherapy.

There are limitations for the current work that we plan to improve in future. 
First, the proposed ME spotting method is our first attempt for solving the very challenging spontaneous ME spotting task. One limitation of the current method is that it uses fixed interval to detect the peak time points, in future works we plan to improve the method for detecting more specifically the onset and offset frames of each ME. 
Secondly, another limitation of the ME spotting method is that non-ME movements like eye blinks have to be ruled out from real ME cases, and one possible solution listed in our future work plan is combining AU detection with FD process. 
Thirdly, the current available ME datasets are still limited considering the data size and the video contents. The SMIC and CASME data only include faces of near-frontal view, so that 3D head rotation problem was not concerned in the current proposed method. New spontaneous ME datasets are needed in the future, with bigger sample size, and more complex and natural emotion inducing environments, \eg interaction or interrogation scenarios where two or more persons are involved. In future work when we consider the ME analysis in such more wild/natural conditions the 3D head rotation problem should be countered in the tracking process.
At last, we also plan to explore using deep learning methods for the ME recognition task.

\appendices

\vspace{-2ex}
\section*{Acknowledgments}
This work was sponsored by the Academy of Finland, Infotech Oulu and Tekes Fidipro program.

\ifCLASSOPTIONcaptionsoff
  \newpage
\fi

\bibliographystyle{IEEEtran}
\vspace{-2ex}
\bibliography{microPAMI_short}

\begin{IEEEbiography}[{\includegraphics[width=1in,height=1.35in,clip,keepaspectratio]{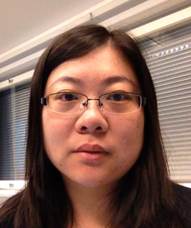}}]{Xiaobai Li}
received her B.Sc degree in Psychology from Peking University, China in 2004, and M.Sc degree in Biophysics from the Chinese Academy of Science, China in 2007. She is currently a
doctoral student in the Center for Machine Vision and Signal Analysis of University of Oulu, Finland. Her research of interests include spontaneous vs. posed facial expression comparison, micro-expression and deceitful
behaviors, and heart rate measurement from facial videos.
\end{IEEEbiography}

\begin{IEEEbiography}[{\includegraphics[width=1in,height=1.25in,clip,keepaspectratio]{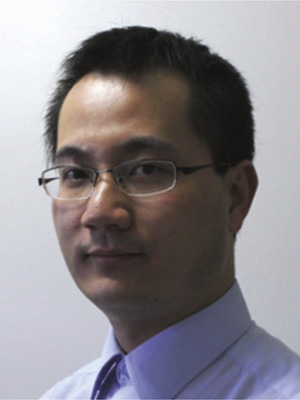}}]{Xiaopeng Hong}
(M'13) received his B.Eng., M.Eng., and Ph.D. degree in computer application from Harbin Institute of Technology, Harbin, P. R. China, in 2004, 2007, and 2010 respectively. He has been a Scientist Researcher in the Center for Machine Vision Research, Department of Computer Science and Engineering, University of Oulu since 2011. He has authored or co-authored more than 20 peer-reviewed articles in journals and conferences, and has served as a Reviewer for several journals and conferences. His current research interests include pose and gaze estimation, texture classification, object detection and tracking, and visual speech recognition.
\end{IEEEbiography}
\begin{IEEEbiography}[{\includegraphics[width=1in,height=1.25in,clip,keepaspectratio]{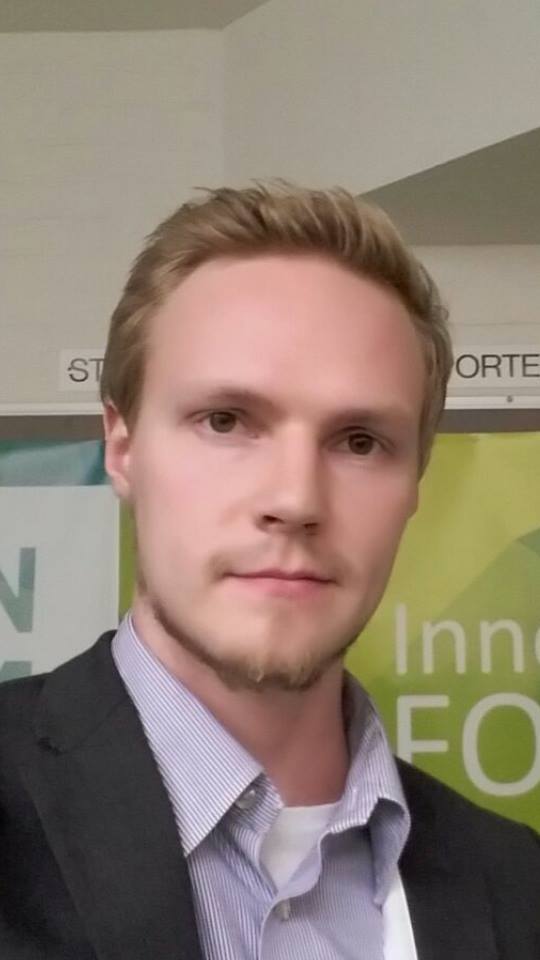}}]{Antti Moilanen}
received his B.Sc. (Tech.) degree in Mechanical Engineering from the University of Oulu, and is currently pursuing the M.Sc. (Tech.) degree in Engineering Physics and Mathematics from the Aalto University. He is currrently working as a Technical Student at the European Organization for Nuclear Research (CERN) in Switzerland.
\end{IEEEbiography}
\begin{IEEEbiography}[{\includegraphics[width=1in,height=1.25in,clip,keepaspectratio]{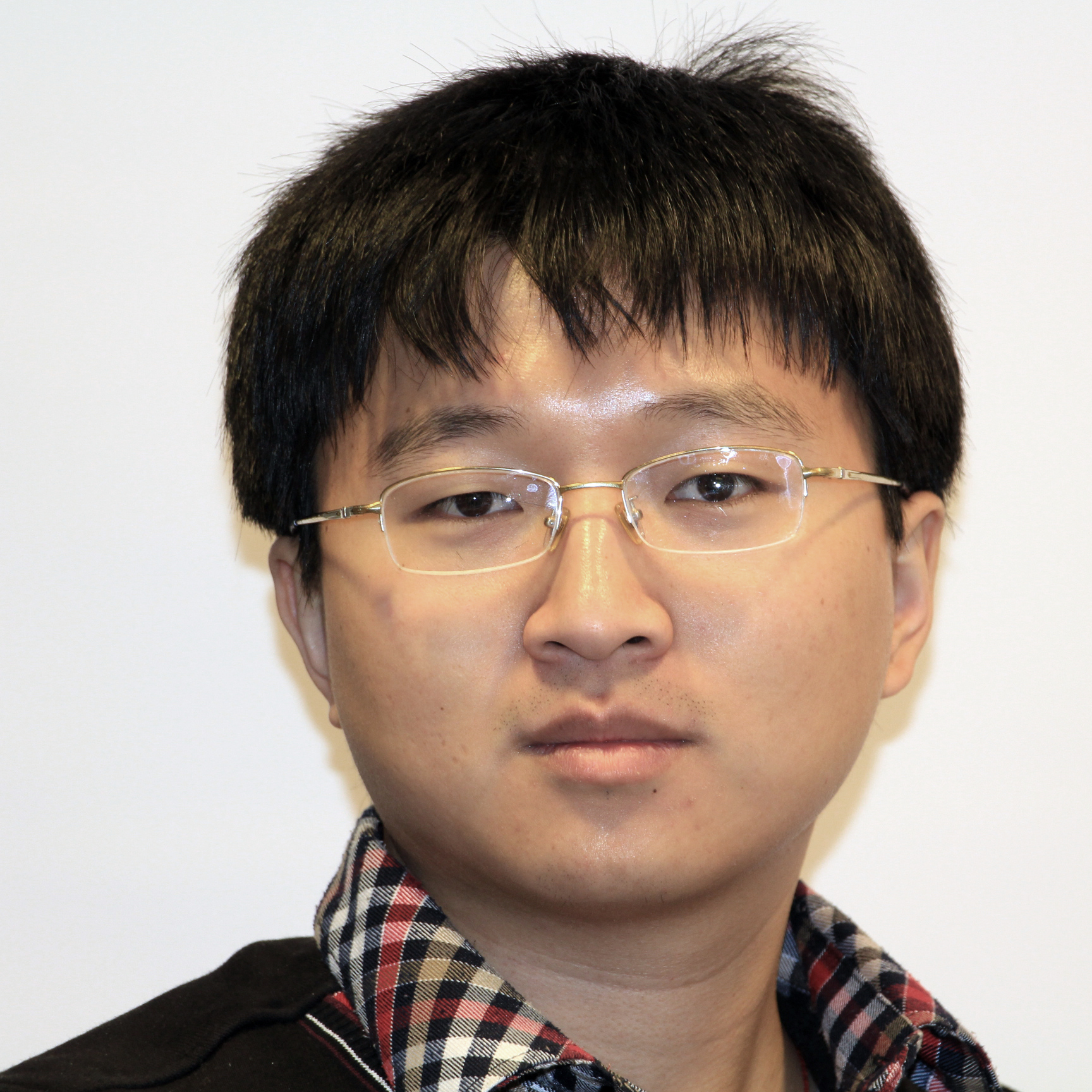}}]{Xiaohua Huang}
received the B.S. degree in communication engineering from Huaqiao University, Quanzhou, China in 2006. He received his Ph.D degree in Computer Science and Engineering from University of Oulu, Oulu, Finland in 2014. He has been a scientist researcher in the Center for Machine Vision and Signal Analysis at University of Oulu since 2015. He has authored or co-authored more than 30 papers in journals and conferences, and has served as a reviewer for journals and conferences. His current research interests include group-level emotion recognition, facial expression recognition, micro-expression analysis, multi-modal emotion recognition and texture classification.
\end{IEEEbiography}
\begin{IEEEbiography}[{\includegraphics[width=1in,height=1.25in,clip,keepaspectratio]{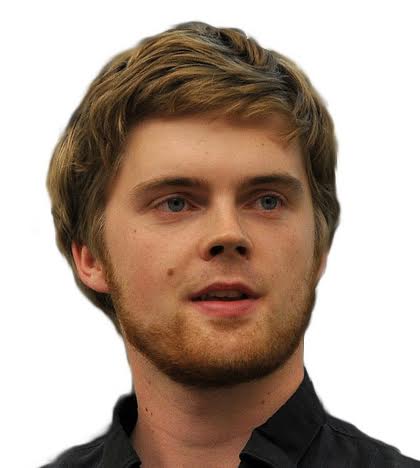}}]{Tomas Pfister}
is a deep learning scientist at Apple. He did his PhD with Prof Andrew Zisserman at University of Oxford on deep learning for computer vision, focusing on automatically understanding human pose, actions and gestures from images and videos. He is the recipient of over 25 research awards, and has publications in most major top computer vision conferences and journals, for which he frequently acts as a reviewer. Before coming to Oxford, he interned at Google in Mountain View working on computer vision for video analysis. He received his BA in Computer Science at University of Cambridge with top grades. His work has received the best paper honourable mention award, the best poster award, and best video award at the British Machine Vision Conference (BMVC).
\end{IEEEbiography}
\begin{IEEEbiography}[{\includegraphics[width=1in,height=1.25in,clip,keepaspectratio]{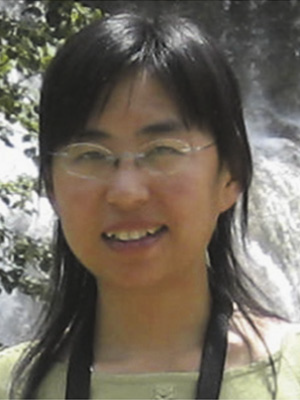}}]{Guoying Zhao}
(SM’12) received the Ph.D. degree in computer science from the Chinese Academy of Sciences, Beijing, China, in 2005. She is currently an Associate Professor with the Center for Machine Vision and Signal Analysis, University of Oulu, Finland, where she has been a senior researcher since 2005. In 2011, she was selected to the highly competitive Academy Research Fellow position. She has authored or co-authored more than 150 papers in journals and conferences. Her papers have currently over 5300 citations in Google Scholar (h-index 32).She has served as area chairs for FG 2017 and WACV 2017 and is associate editor for Pattern Recognition and Image and Vision Computing Journals. She has lectured tutorials at ICPR 2006, ICCV 2009, and SCIA 2013, authored/edited three books and six special issues in journals. Dr. Zhao was a Co-Chair of nine International Workshops at ECCV, ICCV, CVPR and ACCV, and two special sessions at FG13 and FG15. Her current research interests include image and video descriptors, facial-expression and micro-expression recognition, gait analysis, dynamic-texture recognition, human motion analysis, and person identification. Her research has been reported by Finnish TV programs, newspapers and MIT Technology Review.
\end{IEEEbiography}
\begin{IEEEbiography}[{\includegraphics[width=1in,height=1.25in,clip,keepaspectratio]{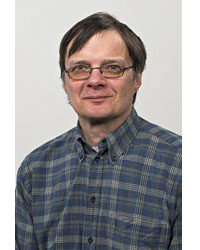}}]{Matti Pietik\"{a}inen}
(F'12) received his Doctor of Science in Technology degree from the University of Oulu, Finland. He is currently a professor, Scientific  Director of Infotech Oulu  and head of Machine Vision Group at the University of Oulu. From 1980 to 1981 and from 1984 to 1985, he visited  the Computer Vision Laboratory at the University of Maryland. He has made  pioneering  contributions, e.g. to local binary pattern (LBP) methodology, texture-based image and video analysis, and facial image analysis. He has authored over  340  refereed papers in international journals, books and conferences.  He was  Associate Editor  of  IEEE Transactions on Pattern Analysis  and Machine Intelligence  and Pattern Recognition journals, and currently serves as Associate Editor of Image and Vision Computing and IEEE Transactions on Forensics and Security journals. He was President of the Pattern Recognition Society of Finland from 1989 to 1992, and was named its Honorary Member in 2014.  From 1989 to 2007 he served as Member of  the Governing Board of  International Association for Pattern Recognition (IAPR), and became one of the founding fellows of the IAPR in 1994. He is IEEE Fellow for contributions to texture and facial image analysis for machine vision. In 2014, his research on LBP-based face description was awarded the Koenderink Prize for Fundamental Contributions in Computer Vision.
\end{IEEEbiography}

\end{document}